\documentclass[sigconf]{acmart}
\usepackage{times}
\usepackage{etoolbox}
\usepackage{todonotes}
\usepackage{pgfplots} 
\usepackage{multirow}
\usepackage{caption}
\usepackage{subcaption}
\usepackage{enumerate}
\usepackage{enumitem}
\usepackage{comment}
\usepackage{hhline}
\usepackage{makecell}
\usepackage{balance}

\newbool{showComments}
\ifbool{showComments}{%
\newcommand{\am}[1]{\textcolor{orange}{\textbf{Alessandro:} #1}}
\newcommand{\cm}[1]{\textcolor{orange}{\textbf{Chulhong: } #1}}
\newcommand{\utku}[1]{\textcolor{blue}{\textbf{Utku: } #1}}
\newcommand{\akhil}[1]{\textcolor{red}{\textbf{Akhil: } #1}}
\newcommand{\fk}[1]{\textcolor{cyan}{\textbf{Fahim: } #1}}
\newcommand{\del}[1]{\textcolor{red}{#1}}
}{
\newcommand{\am}[1]{}
\newcommand{\cm}[1]{}
\newcommand{\utku}[1]{}
\newcommand{\akhil}[1]{}
\newcommand{\fk}[1]{}
\newcommand{\del}[1]{}
}

 \newcommand{\parjump}{\vspace{0.2cm}}

\newcommand{\system}{SensiX++}

\AtBeginDocument{%
  \providecommand\BibTeX{{%
    \normalfont B\kern-0.5em{\scshape i\kern-0.25em b}\kern-0.8em\TeX}}}

\setcopyright{acmcopyright}
\copyrightyear{2018}
\acmYear{2018}
\acmDOI{10.1145/1122445.1122456}

\acmConference[Woodstock '18]{Woodstock '18: ACM Symposium on Neural
  Gaze Detection}{June 03--05, 2018}{Woodstock, NY}
\acmBooktitle{Woodstock '18: ACM Symposium on Neural Gaze Detection,
  June 03--05, 2018, Woodstock, NY}
\acmPrice{15.00}
\acmISBN{978-1-4503-XXXX-X/18/06}

\begin{document}

\title{SensiX++: Bringing MLOPs and Multi-tenant Model Serving to Sensory Edge Devices}

\author{Chulhong Min}
\affiliation{%
  \institution{Nokia Bell Labs}
  \city{Cambridge}
  \country{UK}
}
\email{chulhong.min@nokia-bell-labs.com}

\author{Akhil Mathur}
\affiliation{%
  \institution{Nokia Bell Labs}
  \city{Cambridge}
  \country{UK}
}
\email{akhil.mathur@nokia-bell-labs.com}

\author{Utku G\"{u}nay Acer}
\affiliation{%
  \institution{Nokia Bell Labs}
  \city{Antwerp}
  \country{Belgium}
}
\email{utku_gunay.acer@nokia-bell-labs.com}

\author{Alessandro Montanari}
\affiliation{%
  \institution{Nokia Bell Labs}
  \city{Cambridge}
  \country{UK}
}
\email{alessandro.montanari@nokia-bell-labs.com}

\author{Fahim Kawsar}
\affiliation{%
  \institution{Nokia Bell Labs}
  \city{Cambridge}
  \country{UK}
}
\email{fahim.kawsar@nokia-bell-labs.com}

\renewcommand{\shortauthors}{Min et al.}
\renewcommand\footnotetextcopyrightpermission[1]{}

\pagestyle{plain}

\begin{abstract}

We present \system\ - a multi-tenant runtime for adaptive model execution with integrated MLOps on edge devices, e.g., a camera, a microphone, or IoT sensors. \system\ operates on two fundamental principles - highly modular componentisation to externalise data operations with clear abstractions and document-centric manifestation for system-wide orchestration. First, a data coordinator manages the lifecycle of sensors and serves models with correct data through automated transformations. Next, a resource-aware model server executes multiple models in isolation through model abstraction, pipeline automation and feature sharing. An adaptive scheduler then orchestrates the best-effort executions of multiple models across heterogeneous accelerators, balancing latency and throughput. Finally, microservices with REST APIs serve synthesised model predictions, system statistics, and continuous deployment. Collectively, these components enable \system\ to serve multiple models efficiently with fine-grained control on edge devices while minimising data operation redundancy, managing data and device heterogeneity, reducing resource contention and removing manual MLOps. We benchmark \system\ with ten different vision and acoustics models across various multi-tenant configurations on different edge accelerators (Jetson AGX and Coral TPU) designed for sensory devices. We report on the overall throughput and quantified benefits of various automation components of \system\ and demonstrate its efficacy to significantly reduce operational complexity and lower the effort to deploy, upgrade, reconfigure and serve embedded models on edge devices.

\end{abstract}

\settopmatter{printfolios=true}

\maketitle

\section{Introduction}

The recent emergence of edge accelerators  has radically transformed the analytical capabilities of low-power and low-cost sensory edge devices, such as cameras, microphones, or IoT sensors.  These edge devices can now perform cloud-scale, processor-intensive machine learning (ML) inferences locally and eliminate the need to send a large amount of data to remote data centres - thereby offering benefits in efficiency, speed, availability, and privacy \cite{Satyanarayanan2017, montanari2016understanding}. This possibility has uncovered a new era of affordable AI applications, from personal assistants to recommendation systems to video surveillance available through various edge devices \cite{ha2014towards,ordonez2016deep, kawsar2018earables,liang2019audio,lane2010survey,montanari2018surveying}.


Today, most ML capabilities embedded in the edge devices are limited to inference tasks, i.e., the point at which all the data-driven learning accumulated during training in a data centre is deployed on real-world data to infer a prediction. This task entails  i) preparing the data acquired from a real-world sensor (e.g., pixels from an image sensor or waveforms from a microphone) to a compatible input format, ii) executing the model within a model-specific framework (e.g., TensorFlow or PyTorch) and iii) serving the inferences through well-defined interfaces (e.g., REST APIs). Some applications might demand these devices to store inferences to support historical queries and offer provisioning schemes for deployment. Collectively, these operations represent a subset of  Machine Learning Operations (MLOPs) on edge devices.

We have seen significant efforts from academia and industry to offer toolkits to accelerate the execution of ML models on edge devices \cite{lane2016deepx,liu2018demand,reagen2016minerva,georgiev2016leo,han2016mcdnn,fang2018nestdnn,mathur2017deepeye,montanari2020,min2019closer,min2020sensix}. However, they lack this end-to-end view, for instance, these solutions require expensive and manual intervention to prepare and transform input data, and if done inadequately, the model’s performance suffers in the new operating environment due to data and device heterogeneity \cite{mathur2019mic2mic,blunck2013heterogeneity,mathur2018using}. Furthermore, multi-tenancy support is only available in cloud-scale model serving systems (e.g., \cite{crankshaw2017clipper,tfx}). We envision, by overcoming these challenges, we can unleash a new paradigm of \textit{software-defined-sensors}. For instance, a software-defined camera that can perform multiple and differential interference tasks assisted by automated MLOPs to serve various applications simultaneously or on-demand with zero reliance on distant clouds. Such abilities will transform today's dumb and hard-coded sensory edge devices into dynamic, re-configurable, and intelligent computing platforms.

To this end, we present \system\ -- a multi-tenant model serving system with integrated MLOPs for software-defined sensory edge devices. \system\ reduces operational complexity, minimises redundant data operations, eliminates manual intervention, and achieves two crucial properties:  high inference throughput and low inference latency. \system\ follows a modular system design and applies declarative abstraction principles across its various components enabling different models and respective pipelines to be managed and configured automatically on different accelerators (\S\ref{sec:manifest}). Besides, \system\ externalises system-wide data operations away from model execution enabling multiple models to share identical data transformation and featurisation pipelines.

First, a data coordinator manages the life-cycle of sensors and serves models with correct input data through automated transformations, i.e., resolution scaling, sample scaling, or dynamic encoding (\S\ref{sec:data-coordinator}). 

Next, a resource-aware model server executes multiple models in isolation through model abstraction and pipeline automation (\S\ref{sec:model-server}). This component dynamically constructs model-specific pipeline containers with appropriate frameworks (TensorFlow, PyTorch or TensorFlow Lite) considering available processing resources while meeting latency and throughput constraints. This component also leverages a novel feature sharing functionality that enables multiple models to share a common featurisation pipeline, when possible. The third component of \system\ is a low-overhead scheduler that orchestrates the best-effort executions of multiple models across heterogeneous accelerators, balancing latency and throughput applying multiple optimisation heuristics (\S\ref{sec:scheduler}).  Finally, the fourth component of \system\ encompasses a set of microservices with REST APIs to serve synthesised inferences and  continuous deployment (\S\ref{sec:query}). 

We implemented \system\ in Python and NodeJS and deployed it on Jetson AGX and Coral TPU as representative edge accelerators and added support for widely used embedded ML frameworks, such as TensorFlow, TensorRT, PyTorch etc. Given the dramatic cost reduction of these boards, in conjunction with increasingly cheap compute storage, we expect them to soon feature in commodity scale edge devices, such as a smart camera or a smart speaker. We evaluate \system\ with ten different vision and acoustics models across various multi-tenant configurations and demonstrate that \system\ can offer balanced throughput and latency. We also quantify various components of \system\ to prove its efficacy in reducing operational complexities in serving multiple models. In summary, our contributions in this paper include:

\begin{itemize}[leftmargin=*]
\item A first-of-its-kind framework for multi-tenant model serving with automated MLOPs for sensory edge devices.
\item A declarative abstraction mechanism for system-wide orchestration, automated data operation and pipeline construction for multiple heterogeneous models.
\item A set of novel techniques achieved through adaptive scheduling, feature caching, and shared data operation to reduce and bound latency while maximising throughput.
\end{itemize}
\section{Related Work}

\system\ is designed for serving multiple models with automated MLOps on sensory edge devices. In this section, we review related research in these two areas.

\subsection{Serving Models on Edge Devices}

Unlike model training that happens offline, inference usually serves applications directly and needs to be resource-efficient. Naturally, this requirement has drawn significant attention and many techniques have been proposed including architecture scaling~\cite{bahreini2017efficient,georgiev2016leo, lane2016deepx,montanari2020,montanari2019degradable}, model compression and quantisation~\cite{liu2018demand,han2016mcdnn, han2015deep,Bhattacharya16_DnnSparsification}, pruning~\cite{zhu2017prune,Liu18_RethinkingNetworkPruning,li2016pruning}, accelerator-aware compilation~\cite{reagen2016minerva,antonini2019resource} or model partitioning~\cite{Kang2017,Teerapittayanon2017,Ko2018}. 

While models that run on \system\ can benefit from these techniques to meet latency or accuracy targets, the core model-serving capabilities of \system\ are independent of this aspect. Besides, model efficiency alone is insufficient to meet system-wide performance in a multi-tenant setting. \system\ is designed to make systematic decisions concerning how to run various models with different runtime and performance constraints on edge devices to offer predictable throughput and latency. A few research that comes close to these objectives are NestDNN~\cite{fang2018nestdnn}, HiveMind~\cite{narayanan2018accelerating} and DeepEye~\cite{mathur2017deepeye}. In NestDNN~\cite{fang2018nestdnn}, Fang and his colleagues transform multiple mobile vision models into a single, multi-capacity model consisting of a set of descent models with graceful performance degradation to serve continuous mobile vision applications. In a related effort, although cloud-scale, Narayanan et al. proposed HiveMind~\cite{narayanan2018accelerating} that compiles multiple models into a  single optimised model by performing cross-model operator fusion and sharing I/O across models to optimise GPU parallelisation. In ~\cite{mathur2017deepeye}, Mathur and his colleagues present an execution framework that carefully segregates and schedules different computational layers of multiple models. These research efforts offer an excellent foundation for our work. However, they rely on accessing and modifying model architecture and weights. Instead,  \system\ takes a more practical and model-agnostic execution approach. \system\ treats each model as a black box operating on top of its native runtime framework in a container. These resource-aware containers are then dynamically scheduled with varying heuristics considering system status and individual performance targets. 

There has been considerable prior work in multi-tenant model-serving at a cloud-scale. For instance, Velox~\cite{velox} and Clipper~\cite{crankshaw2017clipper} developed at UC Berkeley utilise a decoupled and layered design with useful abstractions to interpose models on top of different runtime frameworks to build low-latency serving systems. TensorFlow Serving~\cite{tfx} is from Google for serving TensorFlow models, and SageMaker~\cite{sagemaker} from Amazon is a more general-purpose  platform to prepare, build, train, and deploy ML models. These vertically integrated frameworks essentially offer containerised environment with micro-services to execute respective models.\system\ is inspired by these systems, their model abstractions, and scheduling optimisations. However, in contrast to these systems, \system\ supports a wide range of ML models and frameworks, resource aware-scheduling, automated data transformation and pipeline, all at edge-scale.

\subsection{Automated MLOPs on Edge Devices}

MLOPs is a relatively new area of data engineering, and so far, most of the attention is focused on automated model training and primarily driven by the industry due to the challenges in running ML systems in the real world~\cite{Paleyes2020,Sculley2015}. For instance, KubeFlow~\cite{kubeflow} operates on top of Kubernetes and aims to simplify the deployment and orchestration of ML pipelines and workflows. Google's TensorFlow Extended~\cite{tfx} has similar objectives and integrates all components of a ML pipeline to reduce time to production.  Uber's Michelangelo~\cite{michelangelo}  is designed to build and deploy ML services in an internal ML-as-a-service platform. There are several other commercial initiatives such as BentoML~\cite{bentoml} or ElectrifAI~\cite{electrifai} that automate and orchestrate ML workflows to serve in a production environment. All of these platforms essentially aim to reduce the transition time from development to production for ML systems, and in many ways, automates different workflows to simplify and scale these systems. However, these systems operate in a cloud environment, and we are yet to see their entrance to edge devices. \system\ is a first-of-its-kind system for bringing these MLOPs capabilities to edge devices. \system\ borrows many concepts from these systems, for instance, automated data transformation, dynamic pipeline or declarative abstraction to automate system-wide orchestration while contextualising them with careful assessment of resource constraints and performance targets.

\section{SensiX++: Design Principles}

The primary objective of \system\ is to bring cloud-scale MLOPs and multi-tenant model serving to sensory edge devices. To this end, we have designed \system\ with the following four design principles.

\begin{itemize}[leftmargin=*]

\item\textbf{Declarative Abstraction for System Orchestration:} The diversity of machine learning models, underlying frameworks and input variabilities poses a significant challenge towards building a multi-tenant serving solution. We argue that the first step to address this challenge is to devise a declarative abstraction that defines the model and explains its runtime requirement and I/O dynamics. In \system{}, we apply such declarative abstraction in a manifest script that acts as the glue to bring various system components together with systematic orchestration.

\item\textbf{Resource-Aware Model Isolation:} Every machine learning model is unique with implicit assumptions on data distributions and underlying library support. It is essential to keep these dependencies intact while running this model in the real-world to maintain accuracy and robustness. However, this maintenance becomes highly complicated in a multi-tenant setting unless we use appropriate abstractions and isolation. In \system{}, we apply model abstraction and process isolation by running each model in a separate container that embodies all runtime dependencies. We create these containers to run on different available processors to avoid resource contention through careful and periodic assessment of the system load and demands.

\item\textbf{Externalisation of Data Operations:}  Most model serving systems today assume that model-specific input is available. This assumption demands significant manual operations in a multi-tenant production environment and particularly cumbersome for edge devices. In \system{}, we externalise all data operations pertained to a model, including data acquisition, data transformation, featurisation. Furthermore, we leverage our declarative abstraction mechanism to automate and optimise these data operations and simultaneously serve multiple models, reducing operation redundancy by sharing and removing manual interventions.

\item\textbf{Fast and Direct Access for Deployment and Query:} Today's edge devices are essentially all-streaming, dumb data producers. By embedding machine learning capability, we are slowly turning them into intelligent perception units. However, they still rely on the remote clouds for I/O and serving queries to applications. In \system{}, we bring cloud-scale query service directly on the edge device to serve real-time queries and manage model deployment -- thereby offering benefits in speed and availability. \system\ maintains a set of micro-services to operationalise these aspects and leverages its model-serving component for continuous deployment without any downtime.

\end{itemize}

\section{SensiX++: Overview}

We begin by offering an overview of \system\ and its various components. In \S\ref{sec:system-desc}, we cover each of these components, underlying challenges and technical details. 

\begin{figure}[t!]
    \centering
    \includegraphics[width=\linewidth]{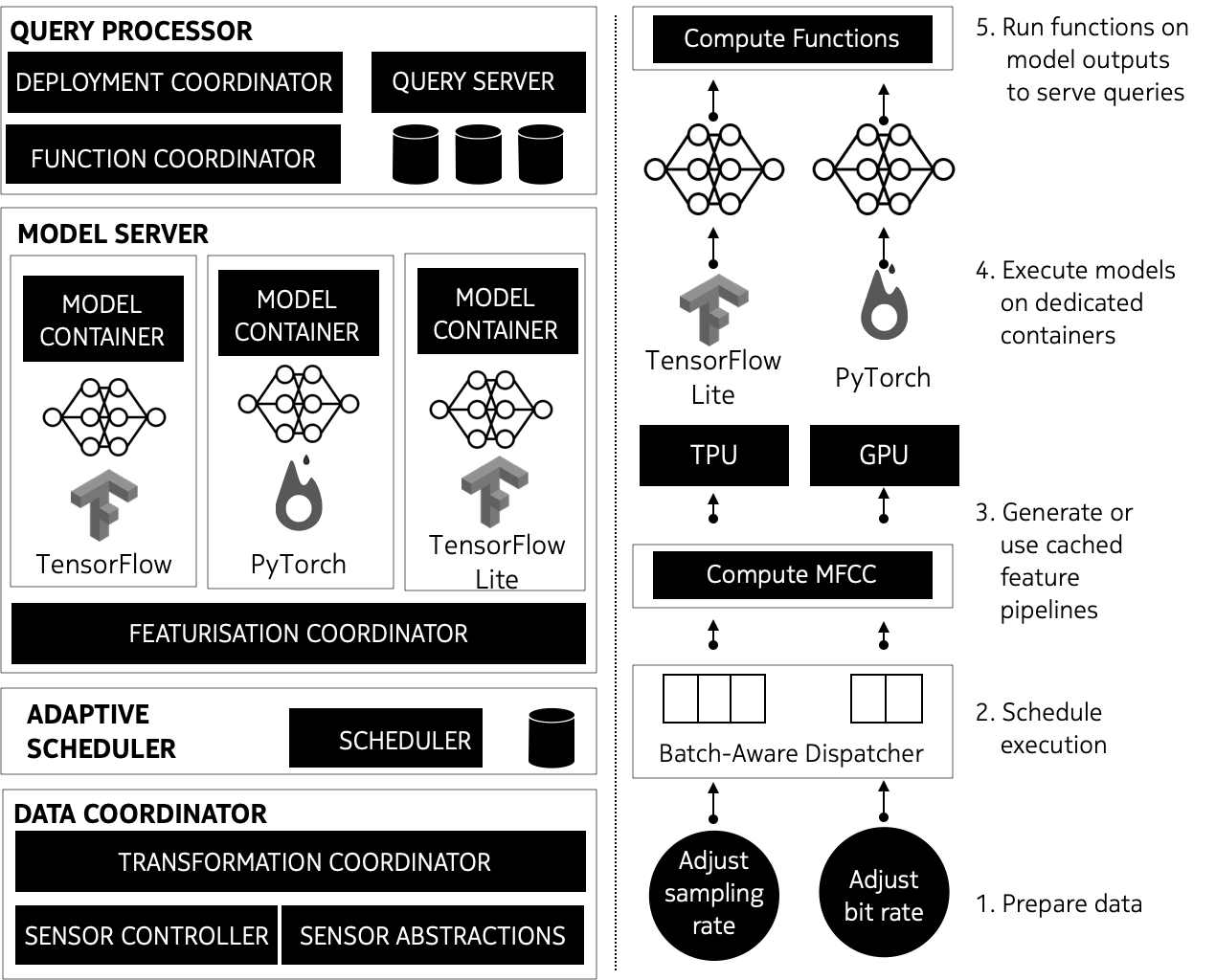}
    \caption{System architecture (left) and an example operational flow for multiple acoustic models (right).}
    \label{fig:operational_flow} 
\end{figure}

\subsection{SensiX++ Components}

Figure~\ref{fig:operational_flow} shows the overall system architecture of \system\ composed of four main components:

\begin{itemize}[leftmargin=*]
\item \textbf{Data Coordinator:} This component manages the sensor life-cycle and reads the sensor data in a unified manner through its sensor controller. It uses a transformation coordinator to transform sensor data to different formats meeting the requirements of different models (\S\ref{sec:data-coordinator}).
\item \textbf{Adaptive Scheduler:} This component is responsible for system-wide orchestration. This component 
decides the execution schedule of multiple models balancing throughput and latency and informs the data requirement of each model, e.g., sensor type, sampling rate, resolution, to data coordinator (\S\ref{sec:scheduler}). 
\item \textbf{Model Server:} This component manages the execution of different models and creates model-specific containers meeting all runtime requirements, i.e., framework and libraries. These containers are then assigned to different processors (CPU, Mobile GPU or TPUs) by the adaptive scheduler. This component maintains a featurisation coordination container that caches feature pipelines to serve multiple models requiring identical features. (\S\ref{sec:model-server}).
\item \textbf{Query Processor:} This component is responsible for external interfaces. It maintains a deployment coordinator to receive model packages and deploy them on \system\ using model server and adaptive scheduler. It also maintains a function coordinator that operates on model outputs to produce synthesised query response served through its query server. Additionally, this component maintains small storage of model outputs to serve historical queries (\S\ref{sec:query_serving}).
\end{itemize}

We illustrate the various MLOPs performed by these components to serve multiple models in Figure~\ref{fig:operational_flow}(right) for a generic acoustic modelling task. Next, we discuss the declarative abstraction mechanism of \system\ that acts as the glue to facilitate the interplay between these components. 
\subsection{Declarative Abstraction for System Orchestration} \label{sec:manifest}

\begin{figure}[t!]
    \centering
    \includegraphics[width=\columnwidth]{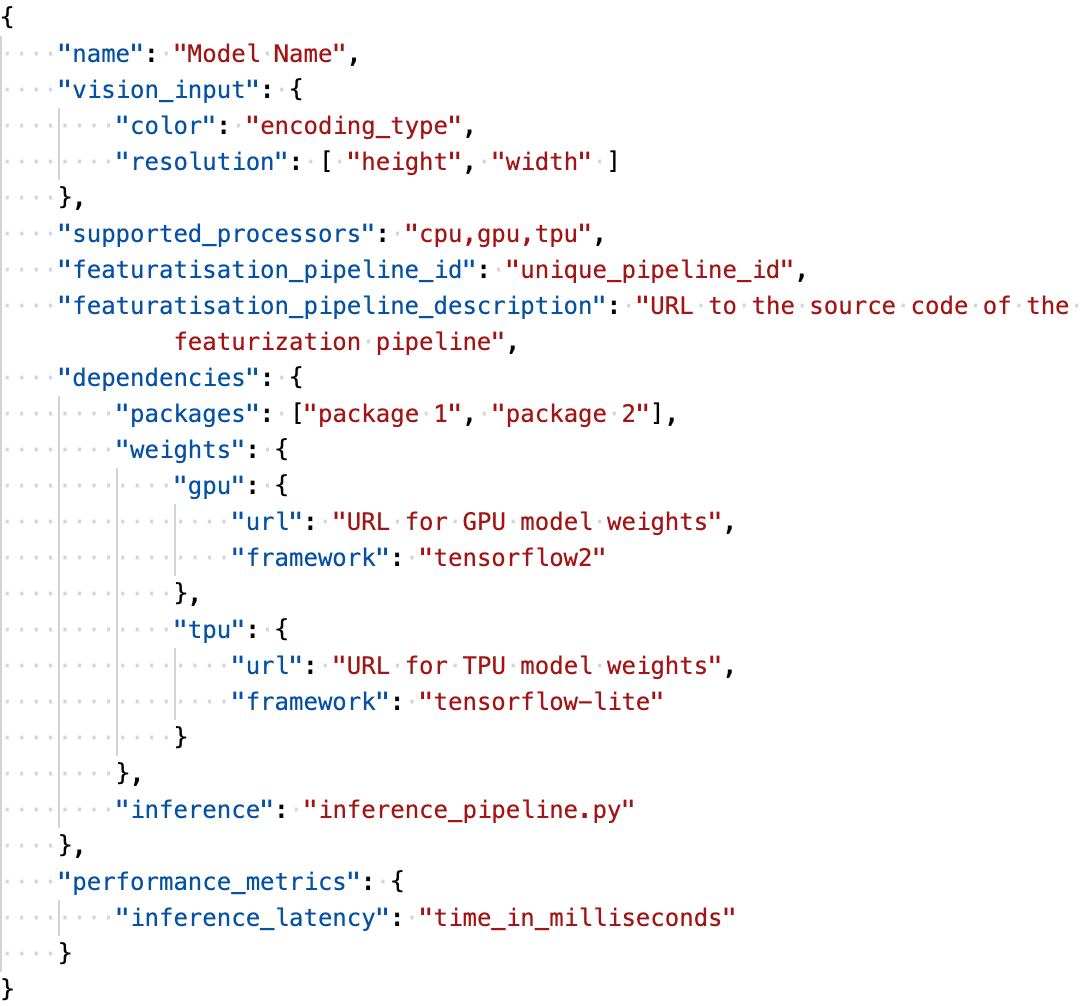}
    \caption{Template of a model manifest file}
    \label{fig:manifest}
\end{figure}

An important design consideration for \system{} is to support efficient execution of ML models on edge devices irrespective of their model architectures, runtime inference frameworks or library dependencies. The challenge in achieving this vision is that a ML model is often released as a black-box and does not contain metadata regarding its input format (e.g., RGB or YUV images), library dependencies, featurisation pipeline etc. Gaining visibility in these aspects of a model is critical to enable many of the system optimisations in \system\ and to achieve significant gains in inference throughput and latency. 

In \system{}, we require model developers to specify metadata about their model through a declarative abstraction mechanism using a \emph{model manifest file}. Figure~\ref{fig:manifest} shows the template of a model manifest file. It contains information about the input requirements of the model (e.g., image resolution and encoding), the inference framework (e.g., TensorFlow 2.0, PyTorch), names of the processors with which the model is compatible (e.g., CPU, GPU, TPU) and a unique global identifier for the feature extraction pipeline (more details in \S\ref{subsec.feature_caching}). Further, the manifest also describes the library dependencies to run the model, a set of processor-specific weights of the neural network model, and an inference script which interprets the numeric outputs of the model (e.g., softmax probabilities) and converts them to human-readable classes). The developers can also specify the latency constraint as a performance requirement (more details in \S\ref{sec:scheduler}).

In the coming sections, we will explain how these meta-information about the model is exploited by various components in \system{} to design novel system-level optimisations. 
\section{SensiX++: Component Description}\label{sec:system-desc}

\subsection{Data Coordinator}\label{sec:data-coordinator}

The lowest layer of {\system} is the Data Coordinator which interfaces with the sensors connected to the host device decoupling data production from data consumption (i.e., ML models) and represents the first aspects of MLOPs to enable the automatic deployment of diverse ML models.

\begin{figure}[t]
    \centering
    \includegraphics[width=0.95\columnwidth]{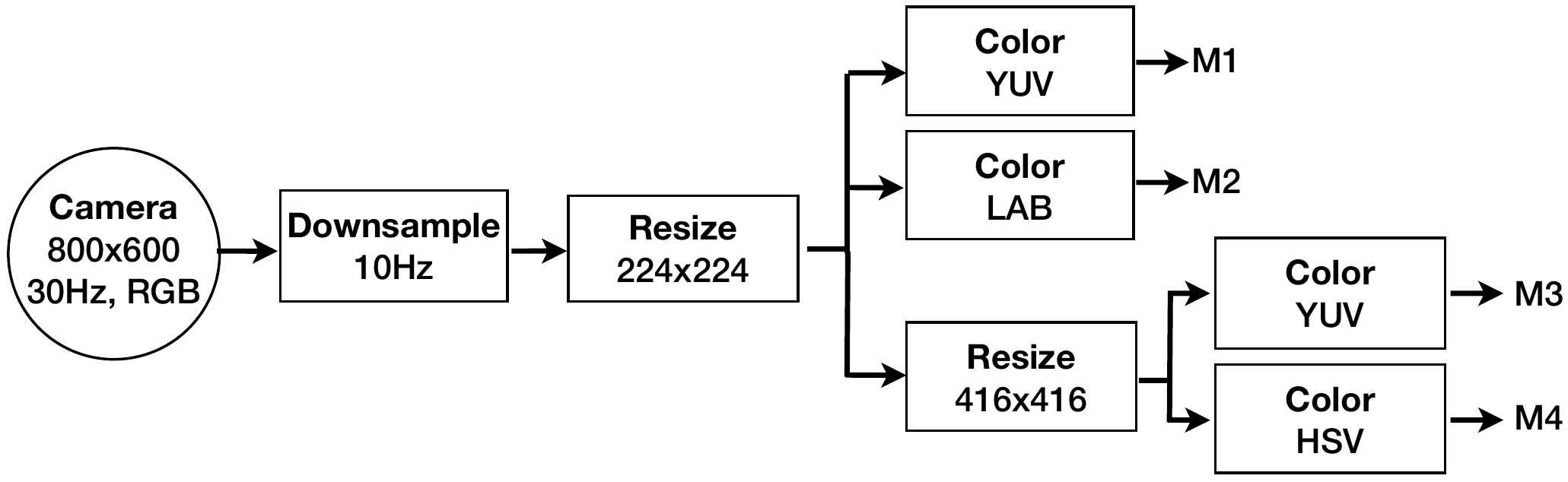}
    \caption{Shared pipeline to serve four models.}
    \label{fig:shared_pipeline}
\end{figure}

At a conceptual level, the data coordinator deals with sensor \textit{capabilities} on one side and model \textit{requirements} on the other. The sensor capabilities are intrinsic characteristics of the sensors. For example for a camera, its capabilities include sampling rate, resolution and colour space (e.g., RGB or YUV). Since {\system} is designed to coordinate the execution of diverse models, not tailored to any specific hardware, the data coordinator layer needs to support models with different sensor data requirements. The model requirements define the expectations each model has on its input data to complete the computation and produce a valid inference result. Any deviation from these expectations will result in the failure of the model to execute or in a severe degradation of its output quality (i.e., recognition accuracy). 

\begin{figure*}
    \centering
    \includegraphics[width=0.9\linewidth]{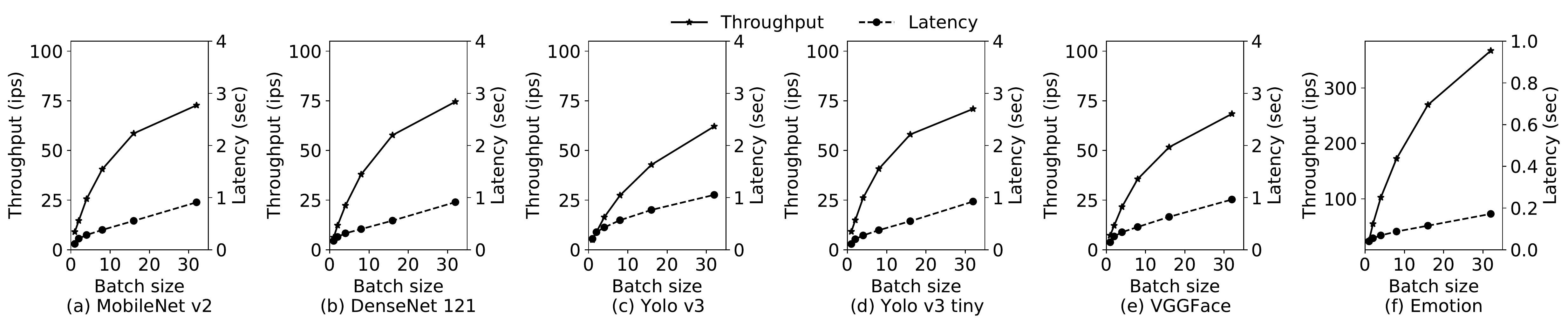}
    \caption{Trade off between throughput (\# of queries per second) and latency (sec).}
    \label{fig:scheduler_batch_tradeoff}
\end{figure*}

In the multi-tenant environment considered in this work it is very likely to have a mismatch between sensor capabilities and model requirements, especially when there is a set of models which needs data from a smaller set of sensors. The challenge for the data coordinator is to resolve the mismatch between sensor capabilities and model requirements automatically and in the most efficient way, ensuring that all the models running are fed with the appropriate sensor data without introducing excessive latency and overhead for the host device. The mismatch resolution also needs to be transparent to the model developers who are not aware of the sensors available on the hosts where their models will be deployed but provide the requirements via the manifest file. 

\parjump{}
\noindent\textbf{Multi-model Sensor Data Transformation.} 
Given the diverse and potentially incompatible requirements of the different models, the sensor-model mismatch cannot be resolved only by re-configuring the underling sensors. Hence, {\system} proposes a two stage approach where the system first carefully configures the sensors to produce data as similar as possible to the model requirements and, second, creates a shared pipeline to take care of the remaining transformations for each model.

\begin{itemize}[leftmargin=*]
\item \textbf{Stage 1}: During the first stage the data coordinator uses the information stored in each model's manifest file to find a common sensor configuration valid for all models. For example, if two vision models require images at 224x224pixels and 416x416pixels respectively, this layer will try to configure the camera to output images at 416x416 resolution, or bigger if that resolution is not available. This avoids sub-optimal sensor configurations which might load the host device unnecessarily given the current models running (e.g., capturing 4k images when models only use 400x400 images).
\item \textbf{Stage 2}: The second stage involves the creation of a pipeline to transform each sensor sample and finally meet the models' requirements. Instead of building an individual pipeline for each model, the data coordinator creates a shared pipeline for all models. This pipeline is manifested as a graph of transformations, an example of which is depicted in Figure~\ref{fig:shared_pipeline}. In this example, four models need to be fed with images from the camera at 10Hz but they all require a different combination of resolution and colour space. By sharing the down-sampling and resize operations across all models the system avoids their repetition in the case when there is an individual pipeline for each model. We show in \S~\ref{sec:benchmarks} how this results in significant savings in terms of CPU and memory load.
\end{itemize}

When building the shared pipeline, the data coordinator places transformations that reduce the number of samples or the size of each sample early in the pipeline (e.g., down-sampling or resize) in order to reduce the operations performed at a later stage. 
Currently, we provide three transformations for cameras and three for microphones, but the data coordinator can easily be extended to include additional ones. For images, {\system} can resize their resolution, modify their colour space or reduce their sampling rate (i.e. frames per second). For microphone, the system can reduce their sampling rate, modify the bit depth of each sample or aggregate samples in different windows (e.g. 1s, 2s).

The components of the data coordinator which implement these functionalities are shown in Figure~\ref{fig:operational_flow}. The Sensor Abstractions hides the details of each sensor and expose a uniform view to the upper layers of the system. The Sensor Controller manages and coordinates multiple instances of sensors and the Transformation Coordinator selects the appropriate sensor configurations and runs the shared pipeline. Collectively, the interplay between these components lead to a completely automated data management for serving models - an integrated part of MLOPs.

In the next section we present the Adaptive Scheduler layer which sits on top of the Data Coordinator and is responsible for the optimisation of the models' execution schedule.

\subsection{Adaptive Scheduler}\label{sec:scheduler}

Due to the dynamic, unpredictable nature of deployment environments, it is infeasible for model serving systems to guarantee the exact performance that models have at development time. This becomes more critical at the edge because it is almost infeasible to dynamically extend the resource availability on the fly. Thus, we design \system\ 1) to allow models to specify their latency requirements, rather than specifying an exact preferred throughput, and 2) to provide \emph{best-effort} performance while meeting those latency budgets.

For a given set of input models, a key to improving their inference throughput without modifying the models is to leverage batch processing. Batch processing enables fast and efficient computation by allowing the internal framework to exploit data-parallel processing, thereby increasing the number of inferences that can be computed per unit of time. However, the downside of batch processing is that it has a higher latency than doing a single inference. Figure~\ref{fig:scheduler_batch_tradeoff} shows the relationships between the batch size, throughput, and latency for different models (details about the models are in Table~\ref{tab:model_list}). Here, we define the latency as the time from the data acquisition to the generation of a final inference output. Interestingly, the model throughput significantly increases even with a small increase in the batch size. For example, MobileNetV2 can be executed 9.1 times per second if the inference is continuously executed with a batch size of 1 (which is a common practice for real-time applications). By increasing the batch size to 4, the throughput is expected to increase to 26.7 inferences per second (about 3 times), but the maximum latency increases only from 0.11 to 0.26 seconds; here, the maximum latency is the latency of the first sample in a batch including the queue time (i.e., the time it takes the data sample to be in the queue). 

\begin{figure}
    \centering
    \includegraphics[width=0.75\linewidth]{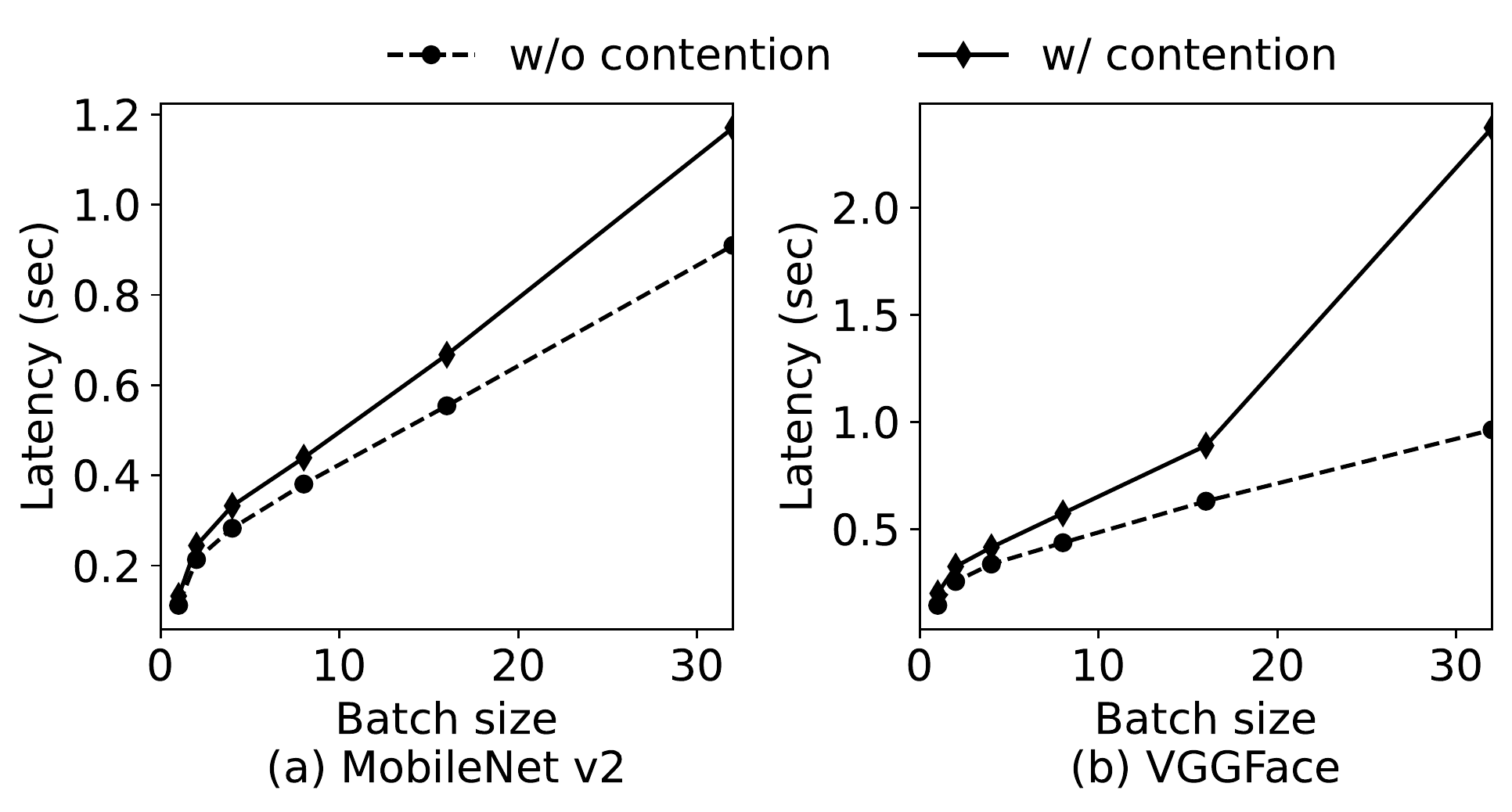}
    \caption{Latency comparison w/ and w/o contention.}
    \label{fig:scheduler_batch_comparison}
\end{figure}

Our explorative study in Figure~\ref{fig:scheduler_batch_tradeoff} indicates an important principle of providing best-effort inference performance. When a model's latency requirement is given in its manifest file, a straightforward method would be to profile the relationship of batch size and latency for the model, and select the maximum batch size for which the expected latency is less than the model's latency requirement. However, such a profile-based static decision does not work well in multi-tenant environments due to resource contention (especially, memory contention) between concurrent models. Figure~\ref{fig:scheduler_batch_comparison} shows latency profiles of MobileNetV2 and VGGFace when they are running alone (w/o resource contention) and together with other two vision models (w/ resource contention). The contention increases the latency, thereby changing the optimal batch size. For example, the optimal batch size of MobileNet v2 for the latency requirement of 300ms changes from 4 (w/o contention) to 2 (w/ contention).

These results have two important implications. First, it may be very costly to profile models by considering the contention effect because all possible combinations of concurrent models need to be addressed. Second, although possible, the profile can still be inaccurate due to the resource contention with other processes and daemons running on the device. 

\parjump{}
\noindent \textbf{Adaptive Scheduling.} To address the aforementioned challenge, we devise a light-weight, adaptive scheduler. The key idea is to monitor the runtime latency of each model and adjust the batch size dynamically. More specifically, the scheduler employs the additive increase/decrease method, inspired by TCP congestion control. Simply speaking, for each model, the scheduler monitors the end-to-end runtime latency of every inference and increases the batch size if the runtime latency is below than the latency requirement. Similarly, it decreases the batch size if the runtime latency exceeds the requirement. Then, based on the selected batch size and runtime latency, the scheduler estimates the expected throughput (as \textit{batch\_size} / \textit{runtime\_latency}) and pass it to the data coordinator as the sampling rate of the corresponding sensor. Then, at runtime, the transformed data in the data coordinator is stored and maintained in the queue of the adaptive scheduler. The batch-aware dispatcher monitors the data queue and triggers the execution of the corresponding model if the number of samples in the queue reaches the batch size. It discards the samples if their queue time exceeds the latency requirement, to avoid unnecessary processing. 

Two practical issues in realising the scheduler are to determine a) how often to change the batch size and b) when to stop increasing the batch size if it is expected to reach the optimal size. First, frequent change would enable the fast saturation to the optimal batch size, but also incur additional cost (e.g., pipeline re-organisation in data coordinator) and make an inaccurate decision due to observation of small samples. Second, a naive implementation of the additive increase/decrease method would repeat to decrease and increase the batch size if it exceeds the optimal one, thereby wasting system resources. 

To address these issues, the scheduler keeps track of the runtime latency and makes an informed decision based on the recent trend, not from a single, last observation. More specifically, it maintains $rl(m, b$), where $rl()$ returns the average runtime latency of recent inferences for given model $m$, and batch size $b$. Then, to make a decision for the batch size of $b$, the scheduler compares the latency requirement with the expected latency of the increased batch size $rl(m, b+1)$, not the current latency ($rl(m, b)$). This enables the scheduler to efficiently spot the optimal batch size while minimising the increasing/decreasing trials.

\subsection{Model Server}\label{sec:model-server}
In this section, we discuss how \system{} enables and optimises the deployment of multiple models on edge devices. 

\parjump
\noindent\textbf{System-aware model containers.}
As the ML ecosystem on edge devices is evolving, there is a wide diversity in the ML frameworks, feature extraction libraries and hardware-specific models. For example, ML models could be developed using different versions of TensorFlow, PyTorch, TensorFlow Lite, may use feature extraction routines provided by OpenCV, Librosa, NumPy, and could be designed to execute on specialised hardware (e.g., TPUs). This challenge of heterogeneity in the ML landscape is further compounded when multiple models run on the same edge device. 

To achieve process isolation and avoid library or framework conflicts between multiple models, \system{} places each model in a separate Docker container. Although such container-based model deployments have been proposed earlier~\cite{crankshaw2017clipper}, \system{}'s key novelty is that the process of creating docker containers and model execution pipelines is completely aware of the system state and is designed keeping multi-tenancy at its core.

In a multi-tenant system, various models compete to access the underlying hardware resources. For instance, due to the benefits associated with GPU acceleration, all model developers may individually want to run their inference pipeline on the GPU, by creating docker containers with GPU support. This can however congest the GPU and require constant paging-in and paging-out of parameters of different models from the GPU memory, leading to higher inference latency and lower inference throughput for each model. 

\system{} addresses this resource contention challenge through a simple yet effective solution. In the model manifest file shown in Figure~\ref{fig:manifest}, model developers specify the hardware-specific variants of their models. For example, for an \emph{Object Detection} task, a developer may provide a MobileNetV2 model (that runs on a GPU), a MobileNetV2-TensorflowLite variant (that runs on a CPU) and a MobileNetV2-TPU variant (that runs on a Coral Edge TPU). At the time of model deployment, the \emph{Container Creator} component first interacts with the \emph{Execution Coordinator} and obtains the current inference workload on each processor (e.g., CPU, GPU, TPU) for the next $k$ seconds. Thereafter, it selects the processor \emph{$P\_min$} with the least inference workload and checks if the model weights provided by the developer are compatible with \emph{$P\_min$}. If yes, \system{} creates a container specific to that processor (e.g., Edge TPU) by downloading the TPU-specific weights and inference pipeline provided by the developer as well as TPU-specific libraries. In case the model developer has not provided a model compatible with \emph{$P\_min$}, \system{} looks for the next available processor in the system, and creates a model container specific to it. 

In effect, our proposed approach of system-aware container creation provides an early-stage load balancing and minimizes the resource contention in a multi-model system right before the models are about to be deployed. 

\noindent
\textbf{Feature Caching.}\label{subsec.feature_caching}
Despite the recent emphasis on end-to-end deep learning on raw data, many ML models still use dedicated feature extraction pipelines to generate features from the raw data and feed them to the ML model, e.g., mel-filterbank generation for audio signals, frequency-domain feature generation for inertial sensor data. Interestingly, different models often use the same feature extraction pipelines before feeding the data to task-specific classifiers. This presents a clear system optimization opportunity by caching the features generated by one model and reusing them for other models, thereby saving redundant feature computations.  

A major challenge here is that ML models do not yet have a standardized way of specifying the feature pipeline used by them. This implies that it is impossible to know if two models share the same pipeline, without doing a thorough code analysis. As a solution, \system{} envisions and proposes that feature pipelines are assigned a unique identifier based on the sequence of operations and parameters used in the pipeline. As an example, the process of generating mel-filterbank features from raw audio data involves decomposing the audio signal in frames, applying Discrete Fourier Transform on each frame to obtain its frequency-domain power spectrum, and re-scaling the power spectrum to the Mel scale. This sequence of steps and the parameters used in each step collectively constitute the feature extraction pipeline. 

In \system{}, model developers can specify a unique identifier to this feature extraction pipeline in the model manifest file. This proposal has parallels with how neural network architectures are assigned unique identifiers; for instance, \emph{MobileNetV2} is simply an identifier for collection of pre-defined computational layers with a fixed set of parameters (e.g., number and size of convolutional kernels, stride length). By adding similar meta-data for feature extraction pipelines, we can facilitate the development of feature caching mechanisms across models which use the same pipeline.

\parjump{}
\noindent
\textbf{Implementation of Feature Caching.} \system{} performs feature extraction and caching in a dedicated docker container known as the \emph{Featurization Coordinator} as shown in Figure~\ref{fig:operational_flow}. When an inference request for a model $X$ is triggered, the Featurization Coordinator receives sensor data from the Execution Coordinator and applies the $X$'s feature extraction pipeline on it. The output features are then fed to $X's$ container for computing the inferences. At the same time, if another model $Y$ registered on \system{} is using the same feature extraction pipeline, these output features computed for $X$ are cached in memory and passed on to $Y$ when it needs to compute an inference on the same data sample(s).  This approach adds a minimal overhead to the inference pipeline associated with caching and retrieval of the output features, however this overhead is negligible in comparison to the gains achieved by skipping the redundant feature extraction.

\subsection{Query Processor}\label{sec:query}

In this section, we discuss how \system\ interfaces with external applications, and handles their queries. 

\parjump{}
\noindent \textbf{Function Coordinator.}\label{sec:postprocessing}
\system\ allows execution of functions that process the outcome from model containers. These functions may serve several purposes including generating higher order of analytics and annotations for the sensor data using  predictions from one or more models or other functions. \system\ dedicates a container in order to facilitate the execution of such post-processing functions applying microservice principles, i.e., a certain function is executed when a particular request arrives. In our case, developers can provide such functions during the deployment phase using two files. A \emph{codelet} file provides a number of executable functions that uses the outcome from one or more other functions and/or models. For each of these functions, a \emph{functions} file lists models and/or other functions, whose outputs are necessary for execution, as well as the type of its outcome. Using this file, we follow an event-triggered approach where execution of a function is prompted by the completion of these entities in the \emph{functions} file. 

\parjump{}
\noindent \textbf{Query Server and Data Store.}\label{sec:query_serving}
To serve query requests from users and applications, \system\ includes a microservice with a number of API endpoints.  \system\ can serve a single response or a stream. The APIs can be summarized as below:
\noindent$\cdot$\verb|/models|: Get the list of models 

\noindent$\cdot$\verb|/functions|: Get the information about available post-processing functions

\noindent$\cdot$\verb|/inference/:type/:function_id|: Get the outcome of a post-processing function with the given query type (single or stream)

In order to serve historical queries, \system\ maintains a LevelDB data store~\cite{leveldb} and the query server leverages this data store to serve queries opportunistically. As the other system components already maintain continuous and fresh pipeline execution from capturing sensor data to post-processing functions and inserting the results into the database, the queries are responded from the data store. This way, even if there are multiple queries with the same request, the response needs to be computed just once.

\begin{figure}
    \centering
    \includegraphics[width=0.95\linewidth]{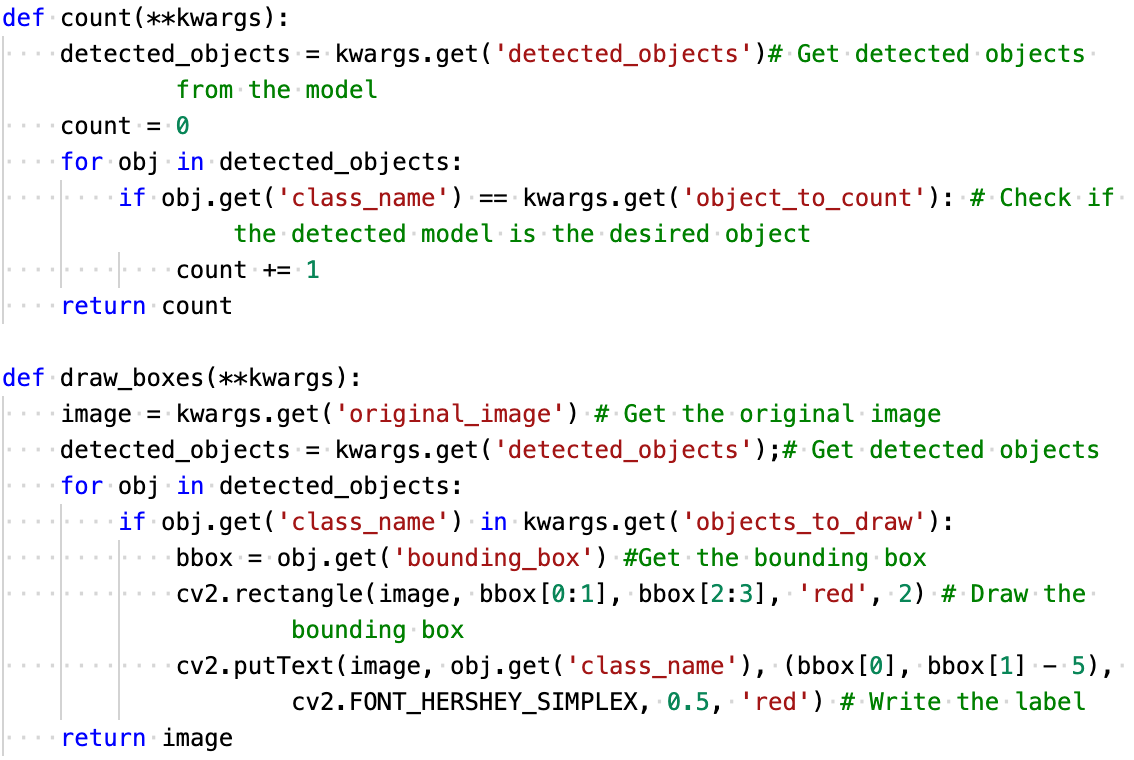}
    \caption{A code snippet for functions in the codelet}
    \label{fig:codelet}
\end{figure}

\parjump{}
\noindent \textbf{Deployment Coordinator.}\label{sec:deployment} \system\ provides a document-based deployment mechanism to model and application developers through an external POST API (\verb|/deploy|). This API accepts an archive of files to define the deployment, composed of a \emph{manifest}, a \emph{codelet} file to add functions to the post-processing container
as shown in Figure~\ref{fig:codelet}, and a \emph{functions} file that is used to create a chain of execution for post-processing functions.  
The manifest file lists a number of model manifests as explained in \S\ref{sec:manifest}. This file is dispatched to the scheduler to create model containers. The codelet and functions files on the other hands are forwarded to the post-processing container to add the new capabilities and to guide pipeline creation from the model output to the query response. 

\section{Evaluation}

In this section, we evaluate \system{} with various vision and audio recognition models. 
First, we investigate the overall inference throughput achieved by \system{} across various multi-model configurations.
Second, we conduct several micro benchmarks to quantify and isolate the benefits of various components in \system{}. Last, we report on an in-depth analysis of the costs and overheads associated with \system{}.

\subsection{Experimental Setup}

\begin{table}[t!]
\footnotesize
\begin{tabular}{|p{0.04\textwidth}|p{0.08\textwidth}|p{0.16\textwidth}|p{0.1\textwidth}| } 
 \hline
  & \textbf{Task} & \textbf{Model} & \textbf{Framework} \\ 
 \hline
 \multirow{7}{*}{\textbf{Vision}} & \multirow{3}{*}{\shortstack[l]{Image\\ classification}} & MobileNet v2~\cite{sandler2019mobilenetv2} & TensorFlow v2 \\ \cline{3-4}
 & & MobileNet v2 (TPU)~\cite{sandler2019mobilenetv2} & Coral TPU \\ \cline{3-4}
 & & DenseNet 121~\cite{huang2018densely} & TensorFlow v2 \\ \cline{2-4}
 & \multirow{3}{*}{\shortstack[l]{Object\\ detection}} & Yolo v3~\cite{redmon2018yolov3} & TensorFlow v2 \\ \cline{3-4}
 & & TinyYolo v3~\cite{redmon2018yolov3} & TensorFlow v2 \\ \cline{3-4}
 & & MobileNet SSD v2~\cite{sandler2019mobilenetv2} & Coral TPU  \\ \cline{2-4}
 & Face\newline recognition & VGGFace (Senet50)~\cite{cao2018vggface2} & TensorFlow v1 \\ \hline
 \multirow{3}{*}{\textbf{Audio}} & Emotion\newline recognition & Emotion~\cite{emotion_pytorch} & PyTorch \\ \cline{2-4}
  & Sound\newline classification & YamNet~\cite{yamnet} & TensorFlow Lite \\ \cline{2-4}
  & Keyword\newline spotting & Res-8~\cite{keyword_spotting} & Coral TPU \\ \hline
\end{tabular}
\caption{List of models}
\label{tab:model_list}
\end{table}

\textbf{Models:} We select a broad range of models mainly tailored for vision and audio recognition tasks, covering diverse types of objectives, frameworks, processors, and model architectures. Table~\ref{tab:model_list} summarises the models.

\parjump{}
\noindent \textbf{Edge accelerators:} For the experiments, we use NVidia Jetson AGX~\footnote{https://developer.nvidia.com/embedded/jetson-agx-xavier-developer-kit} (a GPU-powered edge board released by NVidia) with Google Coral TPU accelerator~\footnote{https://coral.ai/products/accelerator/}. Jetson AGX hosts a 8-core Nvidia Carmel Arm and an 512-core Nvidia VoltaTM GPU with 64 Tensor Cores able to deliver up to 32 TOPs. CPU and GPU share a common bank of 32 GB of LPDDR4 RAM. Google Coral accelerator has Google Edge TPU coprocessor supporting 4 TOPS (int8).

\parjump{}
\noindent \textbf{Sensors:} We use a Wisenet XNV-6080R camera~\footnote{https://www.hanwhasecurity.com/xnv-6080r.html} equipped with microphone to provide images and audio samples.

\subsection{Overall Throughput}~\label{subsec:eval_throughput}

We investigate the overall throughput of \system{} across various multi-tenant configurations.

\begin{table}[t!]
\footnotesize
\begin{center}
\begin{tabular}{|p{0.04\linewidth}|p{0.02\linewidth}|p{0.6\linewidth}|p{0.18\linewidth}|} 
 \hline
 \textbf{Id} & \textbf{\#} & \textbf{Models} & \textbf{Variation}\\ 
 \hhline{|=|=|=|=|}
 $W_1$ & 3 & MobileNet v2, VGGFace, Yolo v3 & Default \\ \hhline{|=|=|=|=|}  
 $W_2$ & 3 & DenseNet 121, TinyYolo v3, Emotion & \multirow{2}{*}{Model} \\ \cline{1-3}
 $W_3$ & 3 & MobileNet v2 (TPU), MobileNet SSD v2,\newline Keyword & \\ \hhline{|=|=|=|=|}
 $W_4$ & 1 & MobileNet v2 & \multirow{3}{*}{\# of models} \\ \cline{1-3}
 $W_5$ & 2 & MobileNet v2, VGGFace & \\ \cline{1-3}
 $W_6$ & 4 & MobileNet v2, VGGFace, Yolo v3, Emotion & \\ \hhline{|=|=|=|=|}
 $W_7$ & 3 & MobileNet v2 (500), VGGFace (500),\newline Yolo v3 (500) & \multirow{2}{0.2\linewidth}{Latency requirement} \\ \cline{1-3}  
 $W_8$ & 3 & MobileNet v2 (300), VGGFace (500),\newline Yolo v3 (1000),  & \\ \hline   
\end{tabular}
\end{center}
\caption{List of workloads; the parenthesized number represents the latency requirement in milliseconds.}
\label{tab:workload_list}
\end{table}

\begin{figure*}[t!]
	\centering
	\begin{subfigure}{0.33\textwidth}
		\centering
		\includegraphics[width=\textwidth]{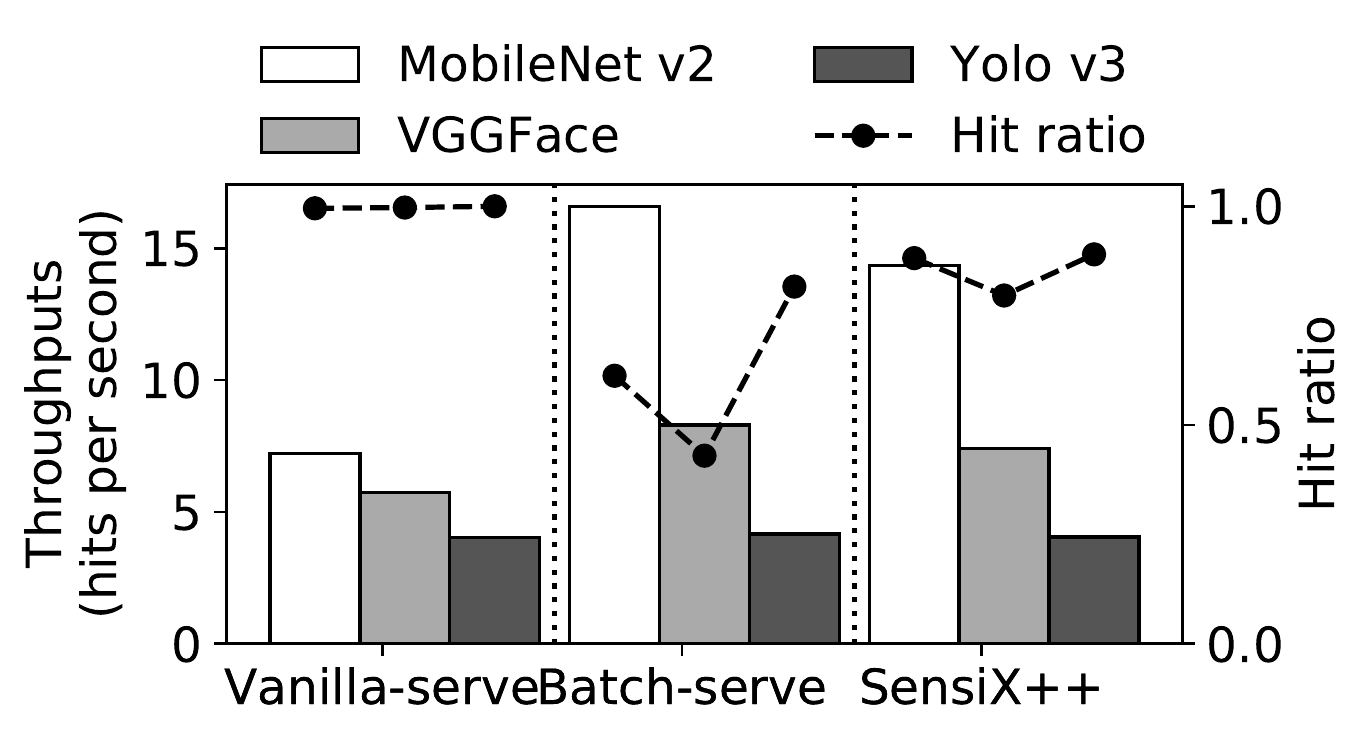}
		\caption{$W_1$ (MobileNet v2, VGG Face, Yolo v3}
		\label{fig:eval_w1}
	\end{subfigure}
    \begin{subfigure}{0.33\textwidth}
    	\centering
    	\includegraphics[width=\textwidth]{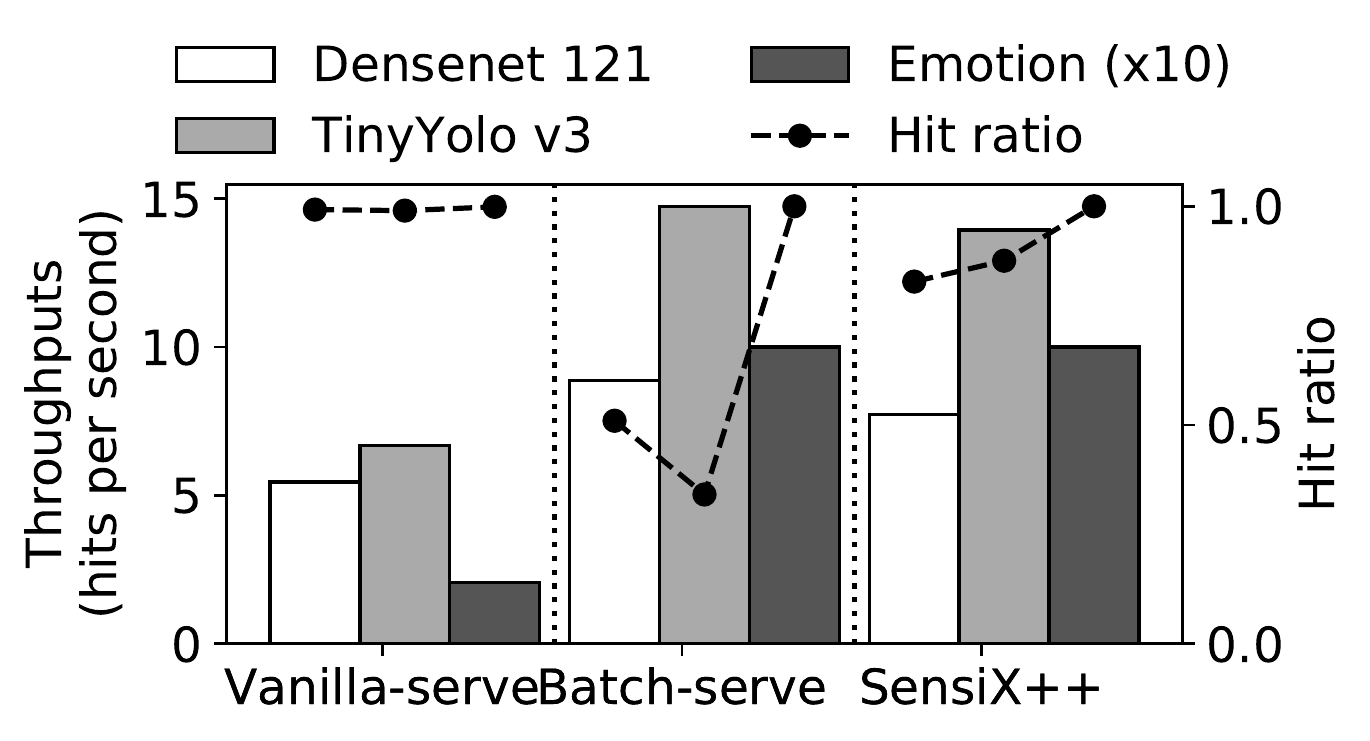}
    	\caption{$W_2$ (DenseNet 121, TinyYolo v3, Emotion)}
    	\label{fig:eval_w2}	
    \end{subfigure}
    \begin{subfigure}{0.33\textwidth}
    	\centering
    	\includegraphics[width=\textwidth]{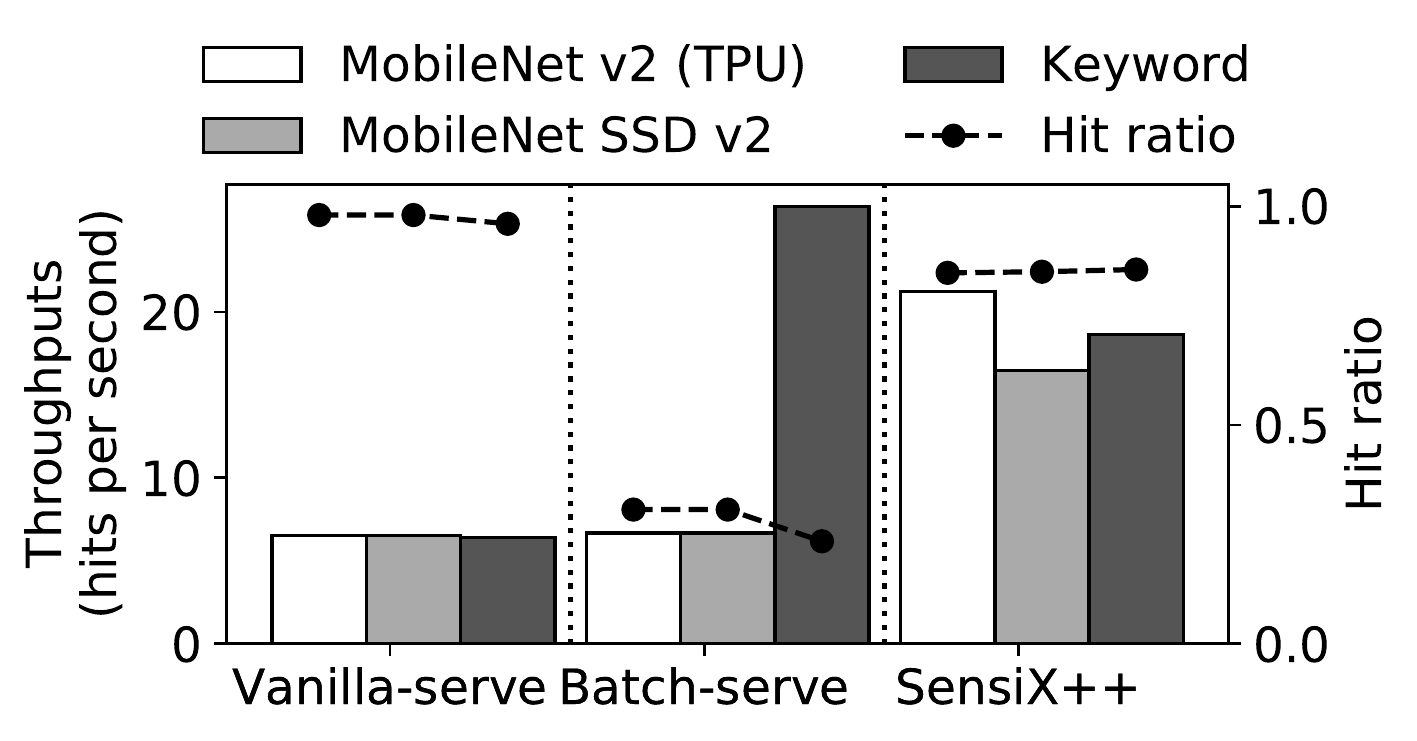}
    	\caption{$W_3$ (Coral TPU framework)}
    	\label{fig:eval_w3}	
    \end{subfigure}
    \caption{Overall throughput across different multi-tenant configurations.}
    \label{fig:eval_overall}
\end{figure*}

\parjump{}
\noindent\textbf{Workloads:} We compose eight workloads to benchmark, by selectively using models in Table~\ref{tab:model_list}. Table~\ref{tab:workload_list} shows the details. We select $W_1$ (MobilNet v2, Yolo v3, VGG Face) as a default workload and vary diverse aspects for the comprehensive analysis in terms of model types ($W_2$, $W_3$), the number of concurrent models ($W_4$, $W_5$, $W_6$), and latency requirements ($W_7$, $W_8$). 
By default, we set 300 ms and 100 Hz to the latency requirements and the maximum inference rate for all models. 

\parjump{}
\noindent \textbf{Baselines:} We compare \system\ against two baselines: 
\begin{itemize}[leftmargin=*]
     \item Vanilla Serving (Vanilla-serve): This baseline is inspired by existing model serving systems such as TensorFlow Serving. For each model, \textit{vanilla-serve} constructs a model container that includes the end-to-end inference pipeline in it (from data capture to model execution). As such, this baseline does not share any data operations with other models nor does it perform batch inferences. 
     \item Static-batch Serving (Batch-serve): This baseline builds over the vanilla-serve baseline by adding batch processing to maximise the inference throughput. However, it does not account for the resource contention issues in a multi-model setting (as discussed in \S\ref{sec:scheduler}). When a model request is added, it profiles the end-to-end latency by varying the batch size (as in Figure~\ref{fig:scheduler_batch_tradeoff}) and finds the maximum batch size that meets the latency requirement. Then, it continuously computes the inference with the selected batch size. 
\end{itemize}

\noindent\textbf{Metrics:} We measure the effectiveness of \system\ by measuring model inference \emph{throughput} that meets the latency requirement of the model as specified in the manifest file. More specifically, we count the number of inferences per second, for which the end-to-end latency is below than the latency requirement of the model. We define the end-to-end latency as the time from data acquisition to model inference. We further measure the efficiency of \system\ by measuring the \emph{hit ratio}, defined as a ratio between a) the number of data samples that succeed in meeting the latency requirement and b) the total number of data samples generated. A higher hit ratio indicates higher resource efficiency, i.e., system resources are optimally used without waste.

\subsubsection{Overall performance} Figure~\ref{fig:eval_overall} shows the model inference throughput of the workloads, $W_1$, $W_2$, $W_3$; the bars and line represent the throughput and hit ratio, respectively. The results show that \system\ achieves higher throughput than Vanilla-serve by virtue of its batch processing and operation sharing. More specifically, \system\ achieves up to 60\% throughput increase in the default workload, $W1$. The throughput of MobileNet v2, Yolo v3, and VGGFace increases from 7.2, 5.7, and 4.0 to 14.4, 7.4, and 4.0, respectively. However, the throughput of Yolo v3 does not increase meaningfully due to its heavy processing load; batch processing of Yolo v3 with the size of 2 already exceeds over the latency requirement, so \system\ does not increase its batch size.

We also compare the performance between \system\ and Batch-serve. Interestingly, both schemes show comparable throughput across the models, but \system\ shows much higher hit ratios, which implies the optimal use of resources. In Batch-serve, the batch size of a model is determined independently to other models, thus the actual throughput the model achieves is lower than the expected one from the selected batch size due to the resource contention. Thus, in $W_1$, the hit ratio of VGGFace of Batch-serve decreases down to 0.43, meaning that more than half data samples generated in a camera are dropped in the queue or the output of the model inferences is discarded due to the violation of the latency requirement. Unlike Batch-serve, \system\ achieves the hit ratio of nearly one even with unpredictable, fluctuating concurrent workloads. We investigate the performance with different combinations of models in $W_2$ and observe the similar trend to $W_1$. Similar to $W_1$, relatively light-weight models (TinyYolo v3 and Emotion) largely benefits from batch processing, e.g., throughput increase of \system{} by up to 2x and 50x compared with Vanilla-serve, respectively; note that we scaled down the throughput of Emotion in Figure~\ref{fig:eval_w2}. 

We further investigate the \system{} performance in Coral TPU framework with $W_3$. Coral TPU framework has different characteristics from other frameworks. First, it does not allow the concurrent access to Coral TPUs from different processors. Thus, we develop a unified container that manages all the inferences of TPU models and adopt a round-robin scheduler for fairness. Second, it does not support batch processing and corresponding internal optimisation in the framework due to the limited memory (8 MB). However, when there are multiple models, the execution of a batch of samples at once in TPU framework also enables high throughput. This is because Coral TPU can afford a very limited number of models in the memory (usually, only one vision model) and needs to write model data every time when a new model is loaded. Thus, the first time the model runs is always slower than the later times and TPU models also show the similar pattern as shown in Figure~\ref{fig:scheduler_batch_tradeoff}.

Figure~\ref{fig:eval_w3} shows that \system\ outperforms the other baselines in TPU framework as well. \system\ achieves the throughput increase by up to around 3x for all models. More specifically, the throughput of MobileNet v2, MobileNet SSD v2, and Keyword increases up from 6.5 (Vanilla-serve) to 21.2, 16.5, and 18.7 (\system{}), respectively. Interestingly, Batch-serve does not increase the throughput of MobileNet v2 and MobileNet SSD. This is because Batch-serve determines the batch size and sampling rate of each model under the assumption that the very model is running alone. However, due to lack of parallel execution capability of TPU framework, the queueing time of data samples becomes longer than expected in order to wait for other models to be completed and many of them are discarded due to the violation of the latency requirement. On the contrary, Batch-serve increases the throughput of the Keyword model by having relatively higher sampling rate, i.e., less number of samples are discarded. However, we can observe that a large portion of data samples are still discarded; the hit ratio of Keyword in Batch-serve is 0.23. 

\begin{figure}[t!]
    \centering
    \includegraphics[width=1\linewidth]{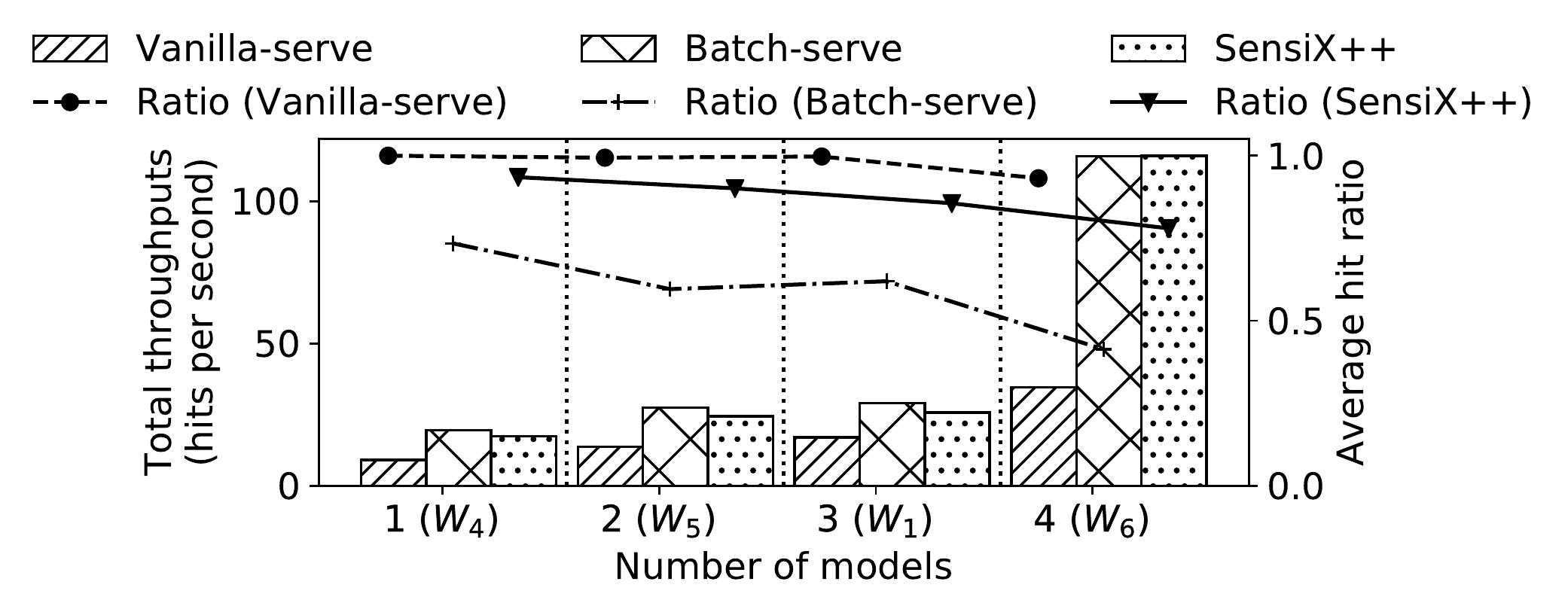}
    \caption{Effect of the number of models.}
\label{fig:eval_n_models}
\end{figure}

\subsubsection{Effect of Number of Models} Figure~\ref{fig:eval_n_models} shows the total throughput and average hit ratio while varying the number of models. The results show that, across different number of models, \system\ achieves higher throughput than Vanilla-serve while maintaining the higher hit ratios than Batch-serve. More specifically, the total throughput increases from 9.1, 13.7, 16.9, and 34.6 (Vanilla-serve) to 17.4, 24.5, 25.8, and 116.0 (\system{}). In the heavy workload ($W_6$), Batch-serve shows comparable throughput, but much lower hit ratio (0.4) than \system\ (0.8). 

\begin{figure}[t!]
    \centering
    \includegraphics[width=\linewidth]{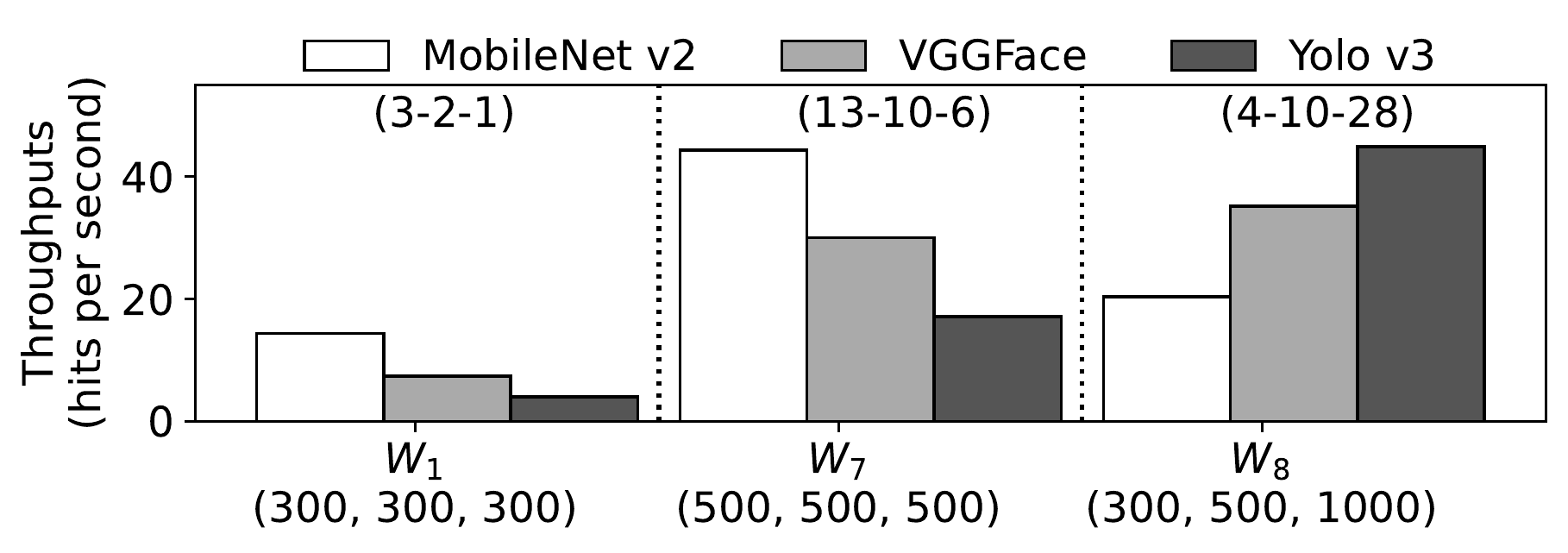}
    \caption{\system{} with different latency requirements.}
    \label{fig:eval_requirement}
\end{figure}

\subsubsection{Performance with different latency requirements~\label{subsec:requirement}} Figure~\ref{fig:eval_requirement} shows the behaviour of \system\ for different latency requirements. The parenthesized numbers in the X-axis is the latency requirement of each model and the numbers in the box represent the batch size which was selected the most during runtime. For example, the most selected batch sizes of MobileNet v2, VGGFace, and Yolo v3 in $W_1$ are 3, 2, and 1, respectively. The results show that \system\ guarantees higher throughput when the loose latency requirement is used. More specifically, when the latency requirement is set to 500 ms in $W_7$, the throughput of three models increases from 14.4, 7.4, and 4.0 ($W_1$) to 44.3, 30.0, and 17.2 ($W_7$). This is enabled by taking the longer batch size from looser latency requirement. For example, the throughput increase of Yolo v3 achieves up to 4x ($W_7$) only at the latency expense of 200 ms. $W_8$ represents the case when concurrent models have different latency requirements, i.e., 300, 500, and 1000 ms. The results show that the adaptive scheduler of \system\ well distributes the resources use based on different requirement. Compared to $W_1$, the throughput increase is different depending on the increase of latency requirement of the model. For example, the throughput increase of VGGFace is 22.6 when the latency requirement is set from 300 ms ($W_1$) to 500 ms ($W_8$), but the increase of Yolo v3 is 40.8 when its latency increases from 300 ms ($W_1$) to 1000 ms ($W_8$). Interestingly, we observe the throughput of MobileNet v2 increases even with the same latency requirement, e.g., 14.4 ($W_1$) to 20.4 ($W_8$). This is because other models use less resources from longer batch size in $W_8$ and MobileNet v2 can be assigned with more available resources.

\subsection{Micro Benchmarks} \label{sec:benchmarks}

We conduct micro benchmarks to quantify the benefits of \system{} components. First, we further break down the performance of \system{}, mainly focusing on the benefits of sharing in two components: a) a shared transformation pipeline in the data coordinator and b) feature caching in the model server. Second, we investigate the performance gain from the system-aware container creation.

\begin{figure*}[t!]
\minipage{0.32\textwidth}
  \includegraphics[width=\linewidth]{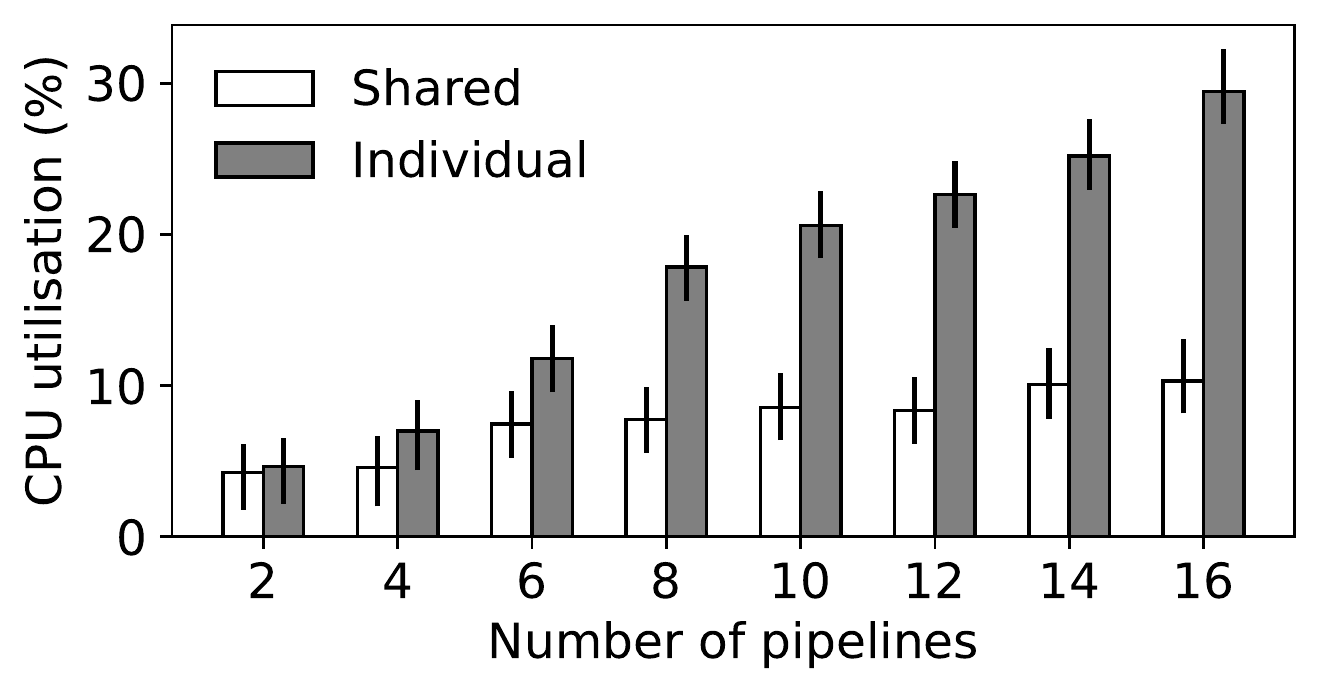}
  \caption{Average CPU utilisation for shared and individual sensor pipelines.}\label{fig:data_coord_cpu}
\endminipage\hfill
\minipage{0.32\textwidth}
  \includegraphics[width=\linewidth]{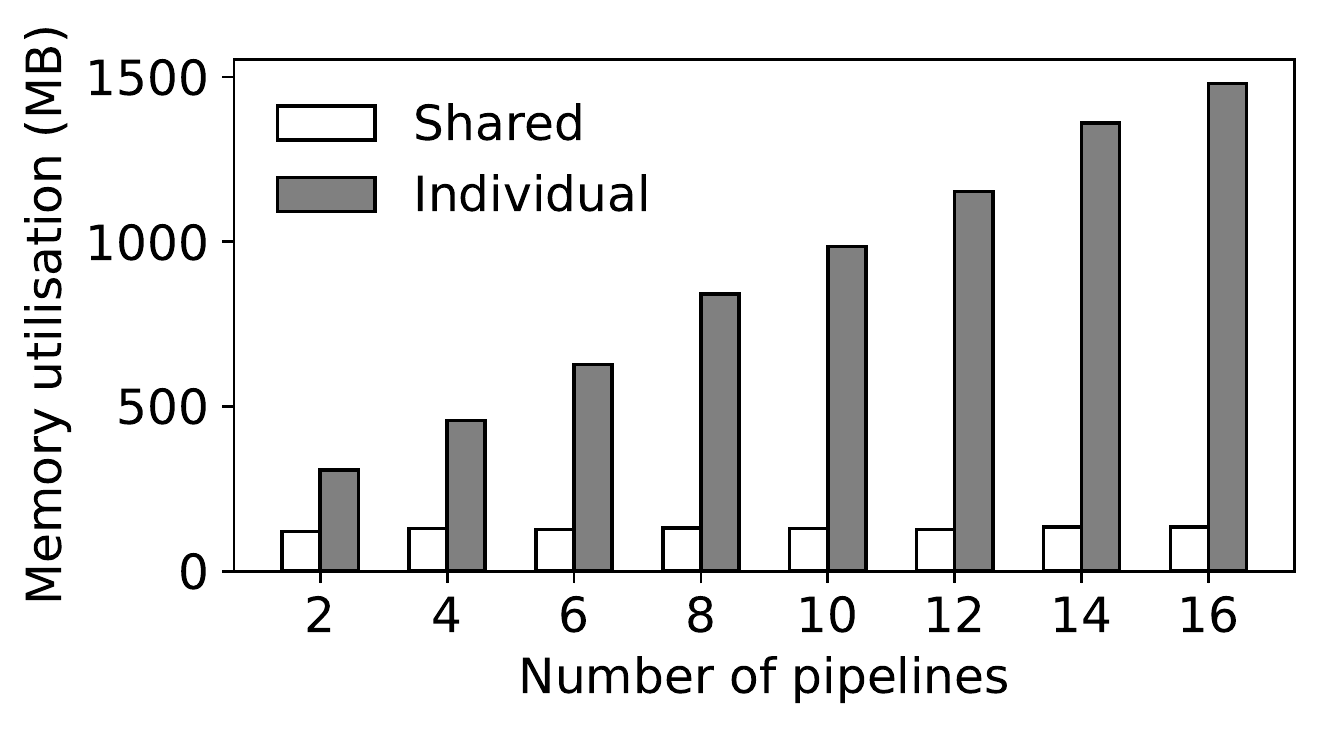}
  \caption{Memory utilisation for shared and individual sensor pipelines.}\label{fig:data_coord_mem}
\endminipage\hfill
\minipage{0.32\textwidth}%
  \includegraphics[width=\linewidth]{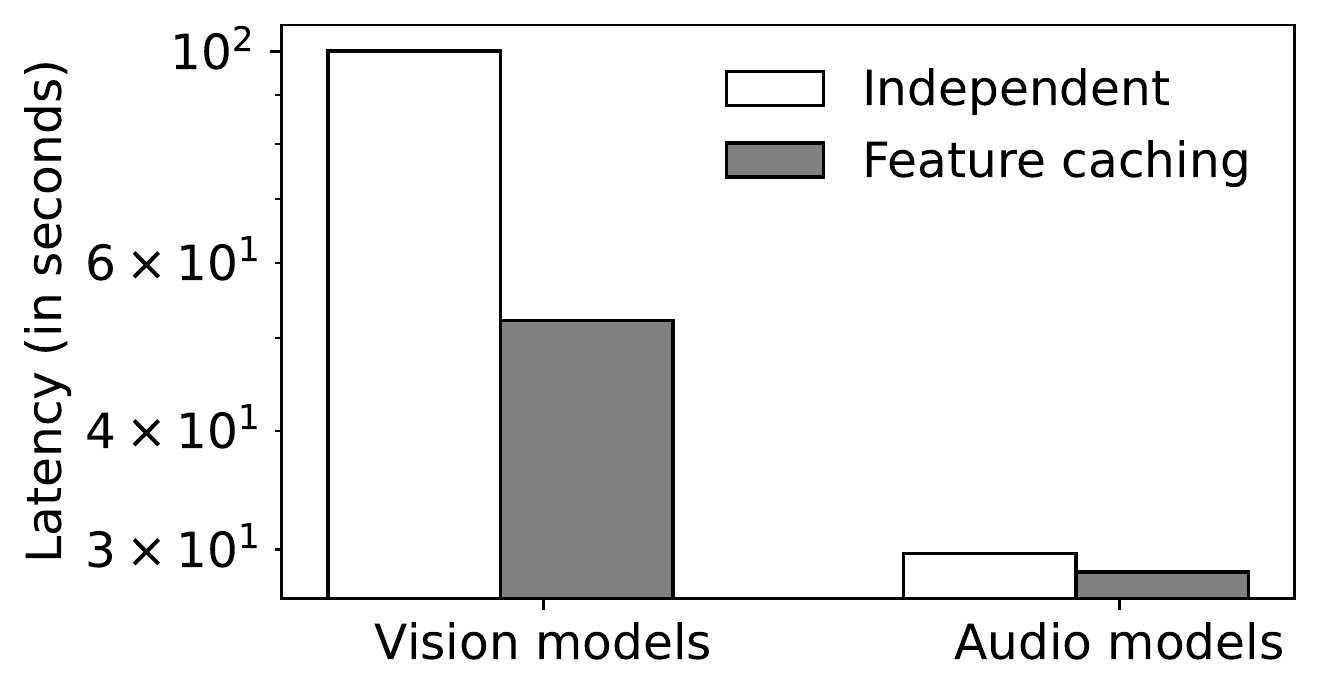}
  \caption{Latency reduction due to feature caching}\label{fig:feature_caching}
\endminipage
\end{figure*}

\subsubsection{Benefits of Shared Transformation Pipeline.} To evaluate the benefit of sharing the transformation pipeline across multiple models we created 16 pipelines inspired from the requirements of the models listed in Table~\ref{tab:model_list}. We deployed these pipelines first using the sharing approach proposed by {\system} (Section~\ref{sec:data-coordinator}) and then as individual pipelines. The latter represents a scenario where each model is deployed individually without awareness of other models running on the system and their requirements, resulting in repetitions in the transformations they perform on the sensor data. Figures~\ref{fig:data_coord_cpu} and \ref{fig:data_coord_mem} report the CPU utilisation and resident set size utilisation of the data coordinator, respectively. We notice that as the number of pipelines increases the benefit of sharing operations results in significant resource savings. The individual pipelines use more resources proportional to the number of pipelines deployed, while the sharing approach scales more slowly with the number of pipelines because operations within the pipelines are executed only once. Already with 4 pipelines deployed we notice significant benefits in the sharing approach which saves 35\% of CPU and 80\% of memory compared to the individual pipelines.

\subsubsection{Latency Reduction using Feature Caching.} To demonstrate the effect of \system{}'s Feature Caching component, we evaluate it on two workloads:
\begin{itemize}[leftmargin=*]
    \item A set of three audio recognition models, namely EmotionNet for emotion recognition, YamNet for acoustic event detection and Res-8 for Keyword Detection. While these models have different inference tasks, they share the same feature extraction pipeline which involves computation of mel-filterbank features from raw audio.  

    \item A set of three visual recognition models, namely MobileNetV2 and DenseNet121 and YoloV3. For these models, we consider \emph{image translation} as the shared featurization operation. Image Translation is a popular mechanism to reduce domain shift (i.e., the divergence between training and test data distributions)~\cite{mathur2019mic2mic}. It involves passing the test image to a pre-trained translation model (e.g., Pix2Pix~\cite{isola2017image}) to obtain a translated image, which is closer to the training data distribution. 
\end{itemize}

Figure~\ref{fig:feature_caching} illustrates the latency reduction achieved by feature caching as compared to the na\"ive baseline when each model independently performs the featurization operations. For the vision models, we obtain a reduction of 48\% in the end-to-end to inference latency. This is primarily because the image translation operation is expensive to perform; hence by doing it once and caching its results for other models results in significant latency improvements. For the audio models, we observe a 5\% latency reduction by sharing the mel-filterbank generation pipeline across models. As this operation is much cheaper than image translation, the latency reduction is smaller. \del{In summary, our results show that by requiring the model developers to provide a trivial piece of metadata about their feature extraction pipelines, \system\ can offer the ability of feature caching and provide clear latency improvements.}

\begin{figure}[t!]
    \centering
    \includegraphics[width=0.85\linewidth]{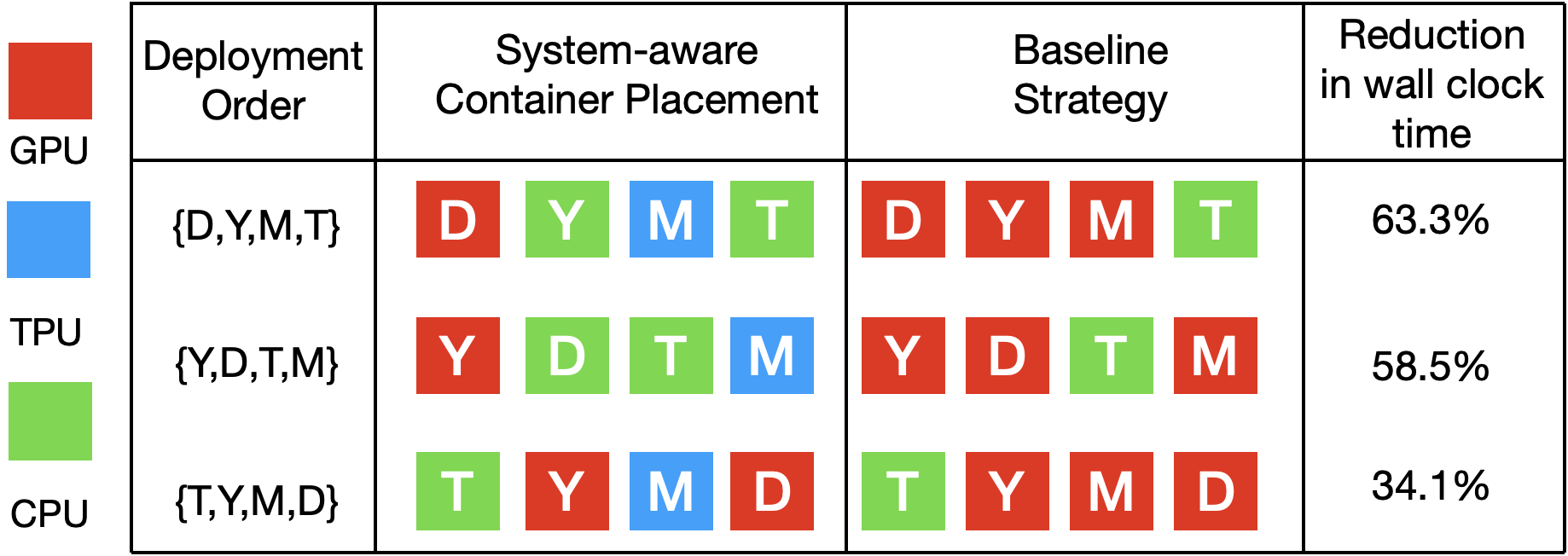}
    \caption{Our system-aware container creation provides a significant reduction in the wall clock time for computing inferences for multiple models.}
    \label{fig:container_creation}
\end{figure}

\subsubsection{Performance Gains from System-Aware Container Creation} We now compare our approach of system-aware creation of model containers against the existing paradigm where model containers are unaware of the system state. Four visual recognition models are used for this experiment, namely \textbf{D}ense121 (CPU or GPU), \textbf{M}obileNetV2 (CPU, GPU or Coral TPU), \textbf{T}inyYoloV3 (CPU), \textbf{Y}oloV3 (CPU or GPU). The values in the parenthesis denote the processors on which a model can be executed. Recall that the processor-specific model weights are specified in the manifest file. 

In Figure~\ref{fig:container_creation}, we show three scenarios of deploying these models. In each scenario, we take a deployment order, e.g., \{\textbf{D,Y,M,T}\} indicates that \textbf{D} is first deployed on the system followed by \textbf{Y}, \textbf{M} and \textbf{T}. When a model is about to be deployed, we check the current resource utilisation of each processor and assign the model to a processor based on its availability and compatibility with the model. For \{\textbf{D,Y,M,T}\}, our algorithm assigns \textbf{D} to the GPU, \textbf{Y} to the CPU, \textbf{M} to the TPU, and \textbf{T} to the CPU. In the absence of this system-aware strategy, a baseline strategy would have assigned \textbf{D}, \textbf{Y} and \textbf{M} to the GPU considering the benefits of GPU acceleration, which would have led to resource contention on the GPU in our multi-model scenario and increased the wall clock time of computing inferences for all the models. Our proposed strategy outperforms the baseline by providing 34-63\% reduction in wall clock time for computing inferences on all the models. 

\subsection{Cost analysis} \label{sec:cost_analysis}

We conduct an in-depth cost analysis to better understand the runtime behaviour of \system{}. To bring MLOps and multi-tenant model serving, besides the operations originally required for the model inference, \system{} additionally performs the following operations: a) the container creation when a model is added, and b) the data transformation and c) adaptive scheduling at runtime, and d) applying post-processing functions and interfacing with a data store in query serving phase. Since the cost of adaptive scheduling is negligible (< 1ms), here, we focus on the other costs.

\begin{figure}[t!]
    \centering
    \includegraphics[width=\linewidth]{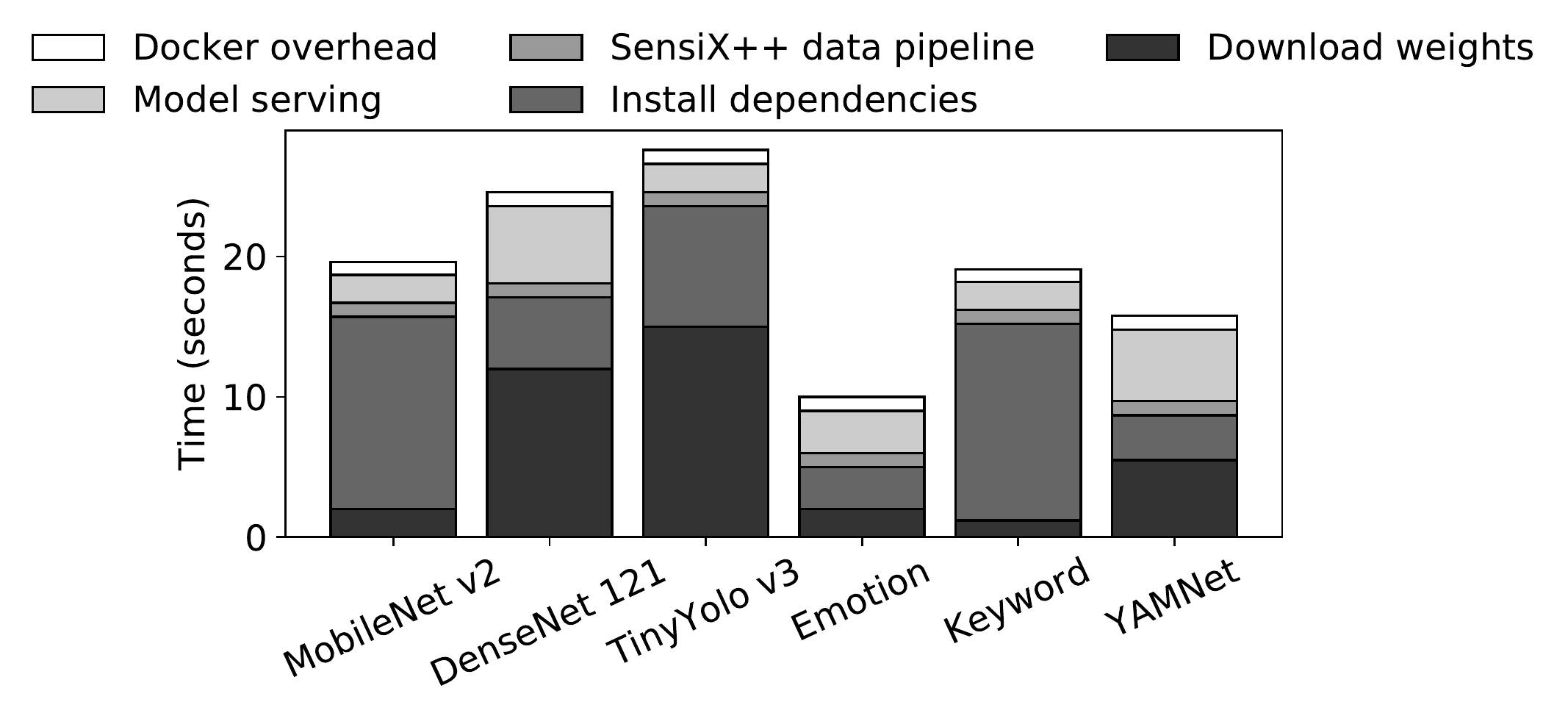}
    \caption{Time taken for creating model containers.}
    \label{fig:docker_creation}
\end{figure}

\parjump{}
\noindent
\textbf{Costs of container creation:} Figure~\ref{fig:docker_creation} shows the time breakdown of various steps involved in creating the docker-based execution pipeline for six different ML models. We observe that the end-to-end deployment of the models in \system{} takes less than 30 seconds. Bulk of this time is spent in downloading the model weights over the network and installing model-specific library dependencies (e.g., for feature extraction) as specified in the model manifest file. \system{}'s overhead -- which includes integrating the \emph{Data coordinator} and \emph{Adaptive scheduler} with the model's inference pipeline and exposing the model interface as a REST API -- is minimal and ranges between 4 to 7 seconds.      
\begin{table}[t!]
\footnotesize
\centering
\begin{tabular}{|l|c|c|}
\hline
                                 & \textbf{Operation} & \textbf{Latency (ms)} \\ \hline
                                 & Abstractions only  & 0.51 (0.11)           \\ \hline
\multirow{3}{*}{\textbf{Vision}} & Sampling rate      & 0.025 (0.053)         \\ \cline{2-3} 
                                 & Resolution         & 1.29 (0.16)           \\ \cline{2-3} 
                                 & Colour space       & 0.30 (0.082)          \\ \hline
\multirow{3}{*}{\textbf{Audio}}  & Aggregation window & 0.024 (0.028)         \\ \cline{2-3} 
                                 & Bit depth          & 7.44 (0.25)           \\ \cline{2-3} 
                                 & Sampling rate      & 9.42 (0.21)           \\ \hline
\end{tabular}
\caption{Average latency of data coordinator operations.}
\label{tab:data_coord_latency}
\end{table}

\parjump{}
\noindent\textbf{Sensor abstractions and transformations latency:} The abstractions and transformations provided by the data coordinator are intended to offer a uniform view of heterogeneous hardware and simplify the deployment of ML models that have not been specifically developed for the sensors in use. As such, these operations should contribute minimally to the overhead of the system and introduce minimal latency in the dispatch of the sensor data to the other components in the system. Table~\ref{tab:data_coord_latency} reports the latency of the individual operations performed by the data coordinator\footnote{We used 640$\times$480 images and audio samples of 1 seocnd at 32 kHz sampling rate.}. We notice that the latency introduced by the abstractions is limited since this is a thin layer over the sensor drivers which transfer data to other components in the system. Similarly, the transformations that reduce the frames per second or aggregate audio samples in different windows do not apply any actual transformation to the data but drop unnecessary frames or aggregate audio sample, hence they introduce very short latency. The other data transformations instead take more time to compute (in particular for acoustic transformations) since they perform modifications to actual samples before dispatching them.   

\begin{figure}[t!]
    \centering
    \includegraphics[width=0.8\linewidth]{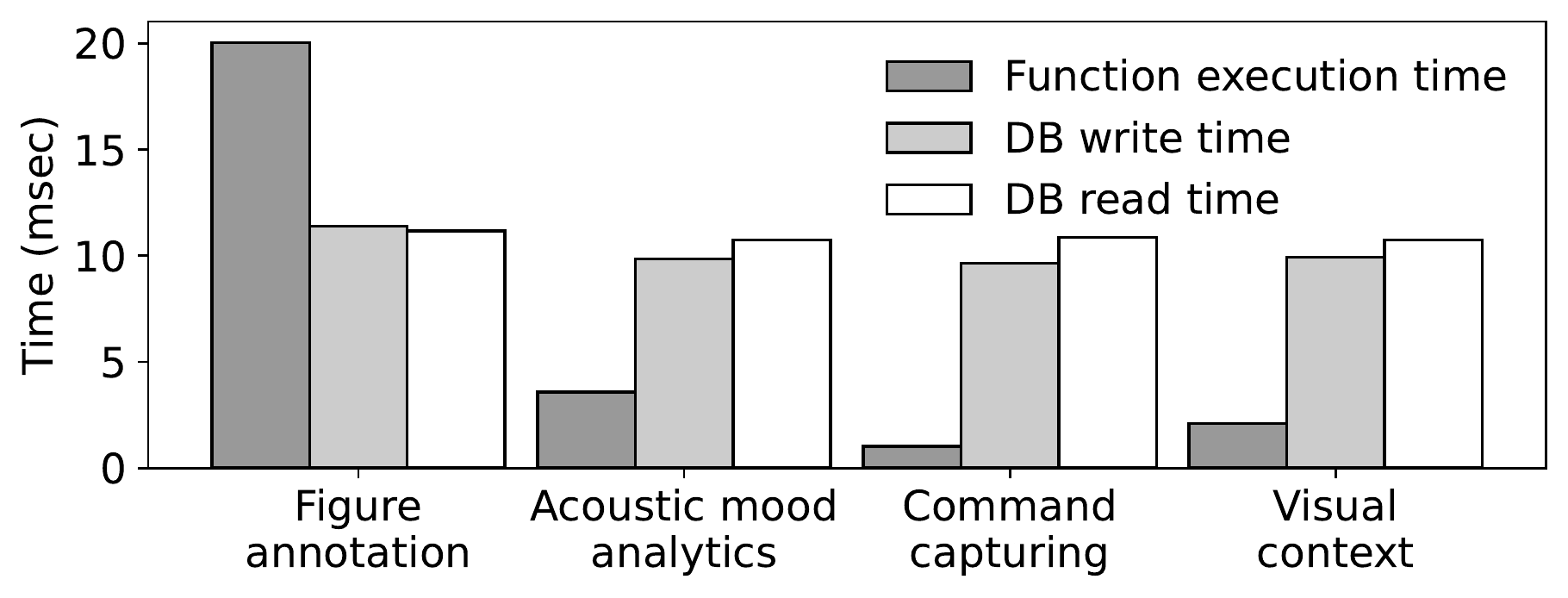}
    \caption{Latency Associated with Query Serving}
    \label{fig:post_processing}
\end{figure}

\parjump{}
\noindent\textbf{Query serving latency:} Figure~\ref{fig:post_processing} shows the latency associated with query serving for four different tasks accumulated by 3 different factors - executing the post-processing function, writing the function output to data-store, and reading from the data store to serve a query. Across these variety of tasks \system\ manages to respond queries within roughly 10ms.

\section{Outlook}

We have presented {\system}, a multi-tenant runtime for model execution with integrated MLOps on edge devices. Thanks to the modular design, {\system} enables great flexibility in the deployment of multiple models efficiently with fine-grained control  on  edge  devices. It minimises data  operation redundancy, managing data and device heterogeneity, reducing resource contention and enabling automatic MLOps. Our benchmarks conducted on an edge device highlight the simplicity of deploying and coordinating diverse vision and acoustics models and the benefits that arise from the automation offered by its key components (i.e., model server, adaptive scheduler and data coordinator). This is a significant step forward compared to current edge MLOps which do not consider multi-tenant scenarios or treat each model as an independent silo without benefiting from system-aware sharing across models and holistic coordination as offered by {\system}.


\balance

\bibliographystyle{ACM-Reference-Format}
\bibliography{reference}


\end{document}